\documentclass[]{interact}

\usepackage[numbers,sort&compress]{natbib}
\bibpunct[, ]{[}{]}{,}{n}{,}{,}
\makeatletter
\def\NAT@def@citea{\def\@citea{\NAT@separator}}
\makeatother

\usepackage{graphicx}
\usepackage{algpseudocode,algorithm}
\usepackage{amsmath,amssymb,amsfonts}
\usepackage{subcaption}

\theoremstyle{plain}

\theoremstyle{definition}

\theoremstyle{remark}

\begin{document}

\jvol{00} \jnum{00} \jyear{2013} \jmonth{January}

\articletype{Full Papers}

\title{
Sparse Representation Learning with Modified q-VAE towards Minimal Realization of World Model
}

\author{
\name{
Taisuke Kobayashi\textsuperscript{a,b}\thanks{CONTACT T. Kobayashi. Email: kobayashi@nii.ac.jp}
 and Ryoma Watanuki\textsuperscript{b}
}
\affil{
\textsuperscript{a}National Institute of Informatics, Japan; and The Graduate University for Advanced Studies (SOKENDAI), Japan
\\
\textsuperscript{b}Division of Information Science, Nara Institute of Science and Technology, Nara, Japan
}
}

\maketitle

\begin{abstract}

Extraction of low-dimensional latent space from high-dimensional observation data is essential to construct a real-time robot controller with a world model on the extracted latent space.
However, there is no established method for tuning the dimension size of the latent space automatically, suffering from finding the necessary and sufficient dimension size, i.e. the minimal realization of the world model.
In this study, we analyze and improve Tsallis-based variational autoencoder (q-VAE), and reveal that, under an appropriate configuration, it always facilitates making the latent space sparse.
Even if the dimension size of the pre-specified latent space is redundant compared to the minimal realization, this sparsification collapses unnecessary dimensions, allowing for easy removal of them.
We experimentally verified the benefits of the sparsification by the proposed method that it can easily find the necessary and sufficient six dimensions for a reaching task with a mobile manipulator that requires a six-dimensional state space.
Moreover, by planning with such a minimal-realization world model learned in the extracted dimensions, the proposed method was able to exert a more optimal action sequence in real-time, reducing the reaching accomplishment time by around 20~\%.

\end{abstract}


\section{Introduction}

Expectations for robots are increasing along with the rapid development of robot and AI technologies, and coupled with the shortage of labor force, robots are beginning to be required to accomplish more complex tasks than ones like factory automation.
For example,
manipulation of flexible objects~\cite{sanchez2018robotic,tsurumine2019deep};
(physical) human-robot interaction~\cite{modares2015optimized,kobayashi2022whole};
and autonomous driving based on high-dimensional observation data from cameras and LiDAR~\cite{paden2016survey,williams2018information}
can be raised.
In these tasks, modeling is a major obstacle to the use of conventional model-based control, where the whole behavior in the task is mathematically modeled in advance for planning the optimal action sequence of the robot~\cite{botev2013cross}.
This is because the state that can adequately represent the whole behavior is unknown and must be extracted from observations somehow.

Recently, a so-called \textit{world model}, which simulates prediction and evaluation of the whole behavior at each time step, has been attracting attention~\cite{ha2018world,hafner2020dream,hafner2021mastering,okada2020planet,okada2021dreaming}.
The world model is acquired from the experienced data under the state, the extraction way of which is also learned from the data, in most cases simultaneously, mainly by a variant of variational autoencoder (VAE)~\cite{kingma2014auto,higgins2017beta,mathieu2019disentangling}.
VAE compresses high-dimensional observation data into a low-dimensional latent space with each axis of the obtained latent space as the state.
The low-dimensional latent space with appropriately compressed observation data can capture the behavior while eliminating unnecessary calculations, and therefore, the world model constructed in this space is suitable for control applications because it enables future predictions accurately with low computational cost.

For the optimal control using the acquired world model, sampling-based nonlinear model predictive control (MPC)~\cite{williams2018information,botev2013cross,okada2020variational} is often employed for its generality.
This is a methodology that randomly generates candidates of the optimal action sequence and finds the better candidates based on the evaluation results of them simulated by the world model with them.
While this methodology can be applied to arbitrary world models because it does not require gradient information, its optimization process is fully accomplished by re-evaluating numerous candidates many times, and requires a very large computational cost.
In particular, this computational cost is correlated to the dimensionality of the state of the world model, and it is intractable to generate the (near) optimal action sequence in real-time if the sufficiently low-dimensional state is not extracted.
On the other hand, of course, if the state dimension is set too small, the whole behavior cannot be simulated by the world model, and the accuracy of the planning itself would be greatly reduced.

Thus, in order to construct a world model that can accomplish the task in real-time, it is essential to keep the size of dimensions of the extracted latent space to a necessary and sufficient level.
In other words, it is desirable to achieve \textit{minimal realization}~\cite{williams2007linear} of the world model.
Most developers to date have adjusted this manually, changing the size of dimensions of the latent space little by little and re-learning to find the minimum size of dimensions that will leave enough information in the state to recover the observation.
Unfortunately, this fine-tuning process is a highly time-consuming process that should be automated.

For this automation, \textit{disentangled representation learning}~\cite{higgins2018towards,kobayashi2020q} (or more directly, the independence and sparsification of the latent space) may play an important role.
In this concept, the latent space should be divided into independent state dimensions and unnecessary state dimensions.
To this end, it requires to eliminate as much as possible the dependencies among dimensions.
In addition, it requires to collapse the state dimensions that can only be dependent so that they are always zero.
If these requirements are always performed correctly, we can hypothesize that the uncollapsed state dimensions correspond to the minimal realization.

In this paper, we focus on one of the latest disentangled representation learning methods, q-VAE~\cite{kobayashi2020q}, for promoting such independence and sparsity.
This q-VAE is derived by replacing the log-likelihood maximization of the observed data, which is the starting point of the conventional VAE, with the $q$-log-likelihood maximization given in Tsallis statistics~\cite{tsallis1988possible,suyari2005law}.
As a characteristic of q-VAE, adaptive learning is performed to balance between the term that improves the reconstruction accuracy of the observed data and the term that refines the latent space, and experimental results have reported that the independence of the latent space is increased.
However, although the cause of this independence could be understood qualitatively, it was not clear whether it was always mathematically valid.
In addition, numerical stability was needed to be guaranteed by ad-hoc constraints.

\begin{figure}[tb]
    \centering
    \includegraphics[keepaspectratio=true,width=0.96\linewidth]{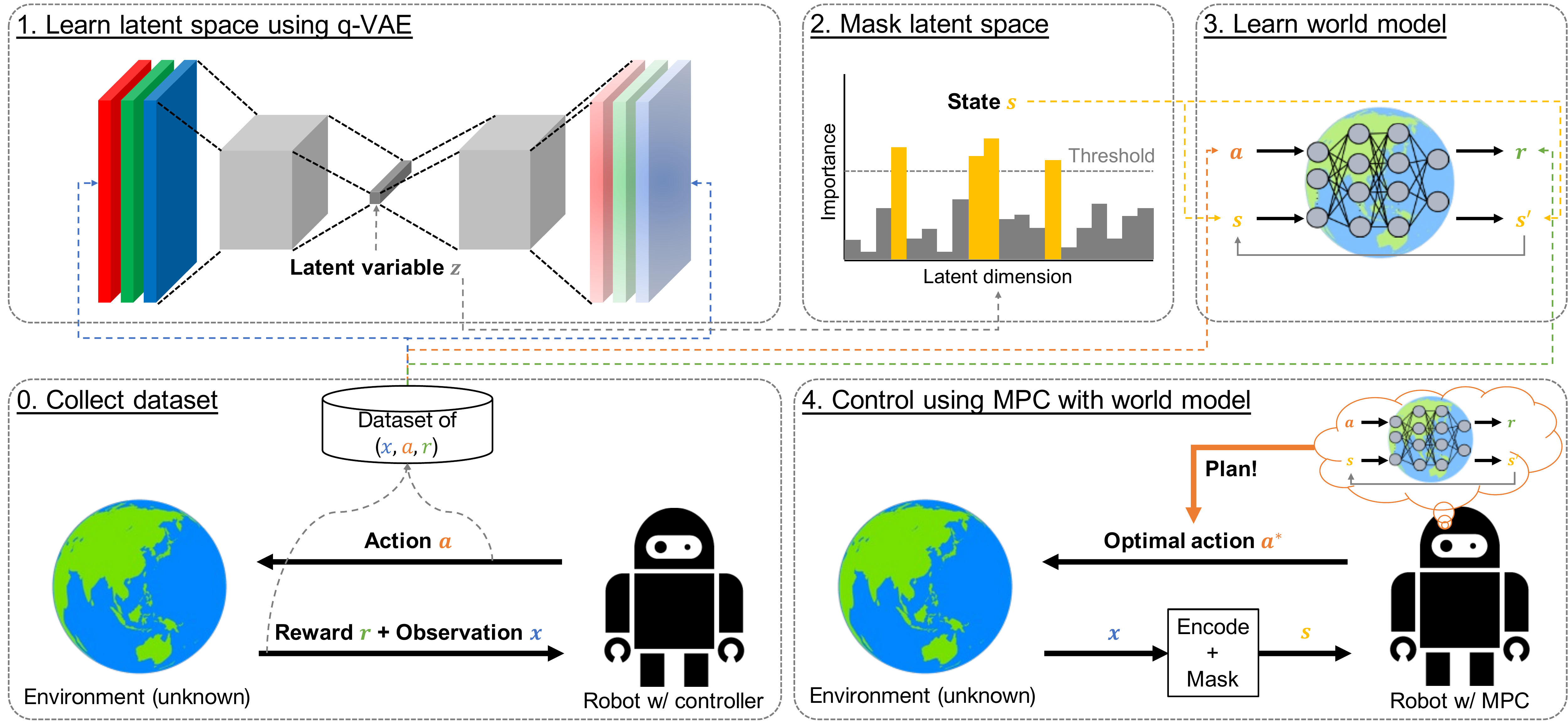}
    \caption{Proposed framework:
        with the collected dataset, the latent space is first extracted using the modified q-VAE;
        since its latent space should be sparse, the state for minimal realization is further extracted by masking the latent space;
        the world model is then trained using the collected dataset and the extracted state;
        finally, the robot plans the optimal action by simulating the future states and rewards using the world model.
    }
    \label{fig:framework}
\end{figure}

Therefore, we deepen the analysis of this q-VAE to establish a new formulation that increases numerical stability and implementation flexibility by eliminating the ad-hoc constraints.
To this end, we exclude common terms that cause instability over all terms found by further decomposing the q-VAE.
We also consider a further lower bound to prevent numerical divergence.
After these modifications, we reveal the conditions to always facilitate sparsification based on the inter-axis dependence and the finite lower bound of the $q$-logarithm.

Using the modified q-VAE, the pre-specified size of dimensions of the latent space can be increased to ensure the reconstruction accuracy of the observed data, and unnecessary state dimensions that can be easily discriminated thanks to sparsification can be masked.
By constructing a world model based on the masked state that may satisfy the minimal realization,
we can expect to accomplish the task in real-time using MPC with the trained world model.
The proposed framework for the above processes is illustrated in Fig.~\ref{fig:framework}.
Note that unlike conventional methods~\cite{hafner2020dream,hafner2021mastering,okada2020planet,okada2021dreaming}, it is not possible to train all neural networks at the same time, but instead, by dividing the optimization problems like~\cite{ha2018world}, the advantages of step-by-step performance analysis and verification can be obtained.

The proposed framework is empirically validated in an autonomous driving simulation and in a reaching task to a target object by a mobile manipulator.
In both tasks, we show that the modified q-VAE improves the sparsity over a conventional method while ensuring the reconstruction accuracy.
We also confirm that the modified q-VAE can make the latent space sparse to six dimensions and achieve almost the minimal realization in the reaching task.
With the world model constructed after masking the unnecessary dimensions, the prediction accuracy can be maintained before the masking.
We finally report that the world model with masking contributes to the improvement of control performance in real-time.

\section{Preliminaries}

\subsection{Model predictive control with world model}

\begin{figure}[tb]
    \centering
    \includegraphics[keepaspectratio=true,width=0.96\linewidth]{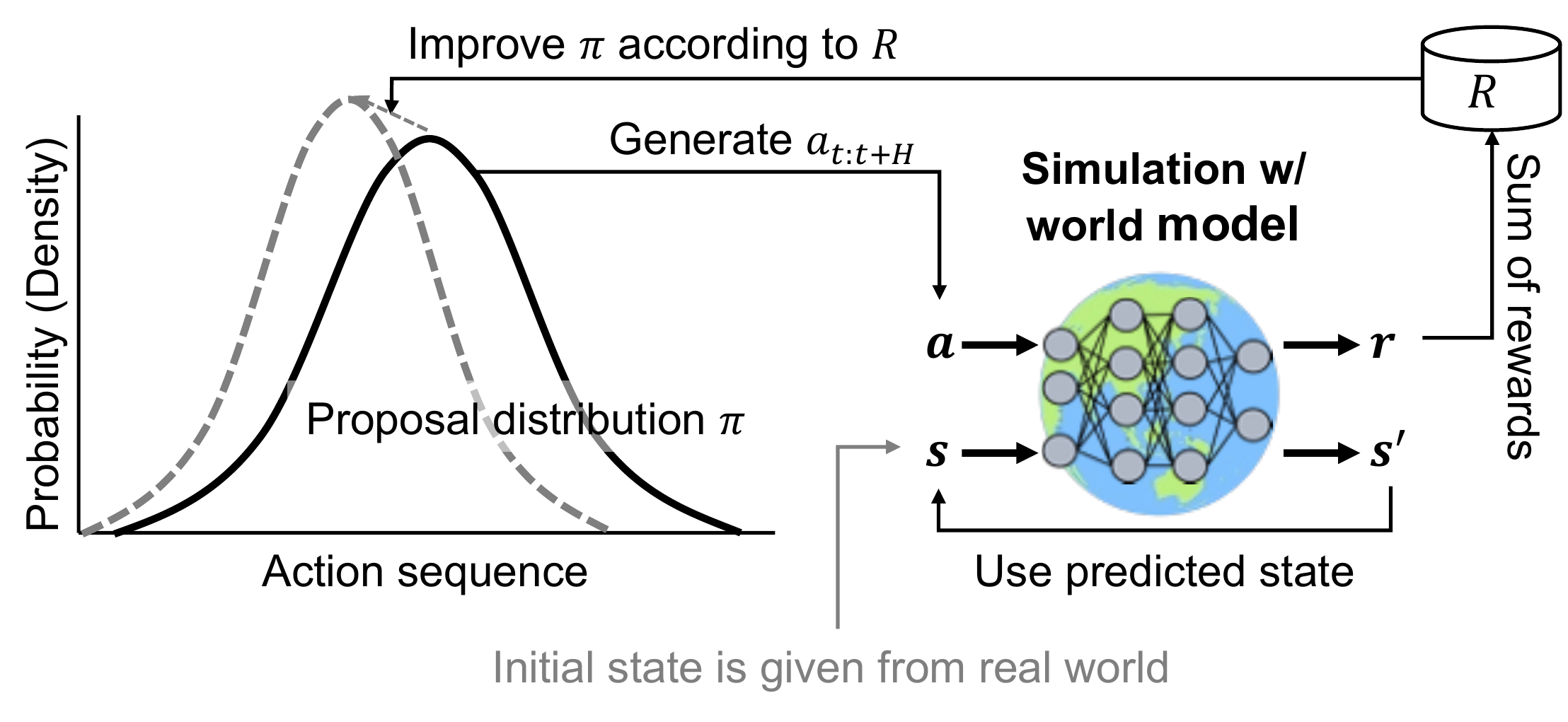}
    \caption{Framework of sampling-based MPC:
        at first, given the current state, the proposal distribution $\pi$ samples the action sequence $a_{t:t+H}$;
        the world model simulates the future states and rewards according to $a_{t:t+H}$ step by step, obtaining the sum of rewards $R$;
        $\pi$ is improved to obtain larger $R$, and repeats the sampling-improvement process until convergence or time limit.
    }
    \label{fig:mpc}
\end{figure}

Before describing the method of extracting the latent space, which is the main topic of this paper, we briefly introduce the world model for the given state and the use of MPC with it~\cite{botev2013cross}.
Here, we first define the state as $s \in \mathcal{S} \subset \mathbb{R}^{|\mathcal{S}|}$, the robot action as $a \in \mathcal{A} \subset \mathbb{R}^{|\mathcal{A}|}$, and the reward (or cost) as $r \in \mathbb{R}$ (with the state and action spaces $\mathcal{S}$ and $\mathcal{A}$, respectively).
Note that $|\cdot|$ with space denotes the size of dimensions of the given space.
In addition, a discrete-time system is often supposed in the world model and MPC.
Therefore, the time step is given as $t = \mathbb{N}$, and it can be noted as a subscript to the above variables to clarify the time of them.

Following the above definitions, we set the world model $\mathcal{W}_{\theta}$ with a set of parameters $\theta$ as follows:
\begin{align}
    \mathcal{W}_{\theta}:
    \begin{cases}
        p_{s}(s^{\prime}_{t} \mid s_{t}, a_{t}; \theta) & \mathrm{dynamics}
        \\
        p_{r}(r_{t} \mid s_{t}, a_{t}; \theta) & \mathrm{reward}
    \end{cases}
\end{align}
where $p_{\cdot}(\cdot \mid \cdot)$ denotes the conditional probability.
That is, with the current state-action pair $(s_{t}, a_{t})$, the world model predicts the future state $s_{t}^{\prime} = s_{t+1}$ and evaluates the current situation as $r_{t}$.
With this structure (i.e. Markov decision process), a transited state $s^{\prime}$ and an evaluated reward $r$ obtained by an action $a$ given to the ``actual'' environment in a state $s$ are combined into a tuple $(s, a, s^{\prime}, r)$, and a dataset with $N$ tuples, $\mathcal{D}_{\mathcal{W}} = \{(s_{i}, a_{i}, s^{\prime}_{i}, r_{i})\}_{i=1}^{N}$, can be constructed and used to train the world model.
Specifically, we can find $\theta \to \theta^{\ast}$ that achieves the following negative log-likelihood minimization problem.
\begin{align}
    \theta^{\ast} = \arg \min_{\theta} \mathbb{E}_{\mathcal{D}_{\mathcal{W}}}[- p_{s}(s^{\prime}_{i} \mid s_{i}, a_{i}; \theta) - p_{r}(r_{i} \mid s_{i}, a_{i}; \theta)]
\end{align}
where $\mathbb{E}_{\mathcal{D}_{\mathcal{W}}}[\cdot]$ denotes the expectation operation by randomly sampling tuples from $\mathcal{D}_{\mathcal{W}}$.

It is important to note that the world model only includes the action as one of the conditions, namely, if the robot freely plans and generates the action sequence $a_{t:t+H} = [a_{t}, a_{t+1}, \ldots, a_{t+H}]$ with $H \in \mathbb{N}$ horizon step, its value can be evaluated via simulating the world model to improve the way to generate $a_{t:t+H}$.
This mechanism is utilized in the sampling-based nonlinear MPC used in this paper, so-called cross entropy method (CEM)~\cite{botev2013cross} (see Fig.~\ref{fig:mpc}).
Based on the evaluation of $a_{t:t+H}$, the optimal $a_{t:t+H}^{\ast}$ is eventually obtained by repeatedly modifying and re-evaluating $a_{t:t+H}$ in the direction of improving the evaluation.
Note that although MPC optimizes the whole action sequence, only $a_{t}^{\ast}$ is actually used, since this optimization is conducted at every time step.

Specifically, CEM samples $K$ candidates of $a_{t:t+H}$, $\{a_{t:t+H}^{k}\}_{k=1}^{K}$, from a proposal distribution (or policy), $\pi$, at each iteration (i.e. the evaluation and improvement), and evaluates all of them using the world model.
The score of each candidate is given as the sum of rewards $R^{k} = \sum_{h=0}^{H} r_{t+h}$.
With this score, $K$ candidates is sorted in descending (or ascending if cost is used instead of reward) order, and then the top $\nu K$ ($\nu \in (0, 1)$ denotes the elite ratio) candidates is extracted as the elites.
Since these elites should be actively sampled, a new policy $\pi^{\prime}$ is obtained through the following maximum likelihood estimation.
\begin{align}
    \pi^{\prime} = \arg \max_{\pi} \sum_{k=1}^{K} \mathbb{I}(R^{k} \geq R_{\mathrm{threshold}}) \ln \pi(a_{t:t+H}^k)
\end{align}
where $R_{\mathrm{threshold}}$ denotes the minimum score in the elites.
$\mathbb{I}(\cdot)$ is defined as the indicator function, which returns one if the condition in the bracket is satisfied; otherwise zero.
If $\pi$ is modeled as normal distribution with $\mu$ location and $\sigma$ scale, this can be analytically solved by the mean and standard deviation of the elites, respectively.

Note that this improvement $\pi = \pi^{\prime}$ is largely sample-dependent, hence if the samples are biased, $\pi^{\prime}$ would overfit to one of the local optima.
To mitigate this issue, the following smooth update is often employed.
\begin{align}
    \theta_{\pi} \gets \eta \theta_{\pi} + (1 - \eta) \theta_{\pi^{\prime}}
\end{align}
where $\theta_{\pi}$ denotes the set of parameters in $\pi$ (in the case of normal distribution, $\theta_{\pi} = [\mu, \sigma]$).
Larger $\eta \in (0, 1)$ makes the update smoother.
The above process (with sampling, evaluation, and improvement) are iterated until the specified number of times or the specified time is exceeded, and the mean of the final $\pi$ updated, or the one with the highest score among the candidates sampled so far, is returned as the optimal action sequence.

\subsection{Tsallis statistics}

\begin{figure}[tb]
    \centering
    \includegraphics[keepaspectratio=true,width=0.96\linewidth]{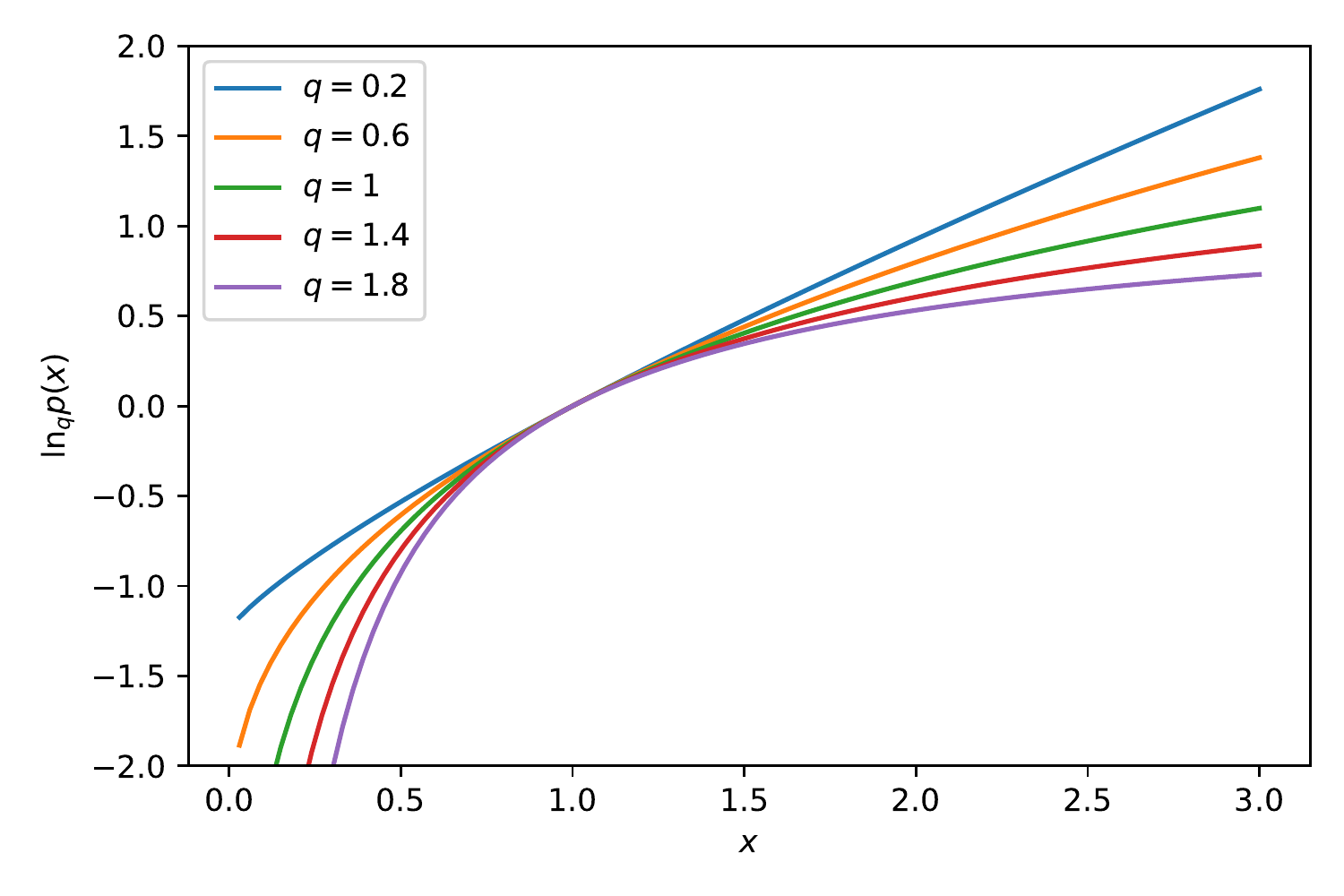}
    \caption{$q$-logarithm:
        with $q > 0$, this function is concave;
        while the output is 0 when $x=1$ for any $q$, the smaller $q$ is, the larger the output.
    }
    \label{fig:qlog}
\end{figure}

Let us briefly introduce several important properties of Tsallis statistics~\cite{tsallis1988possible,suyari2005law}, which is utilized in this paper.
First of all, well-known natural logarithm $\ln(\cdot)$ is extended to $q$-logarithm $\ln_{q}(\cdot)$ with $q \in \mathbb{R}$ in Tsallis statistics.
\begin{align}
    \ln_{q}(x) =
    \begin{cases}
        \ln(x) & q = 1
        \\
        \frac{x^{1-q} - 1}{1 - q} & q \neq 1
    \end{cases}
    \label{eq:qlog}
\end{align}
where $x > 0$.
As illustrated in Fig.~\ref{fig:qlog}, $q$-logarithm with $q > 0$ is concave.
While natural logarithm has infinite upper and lower bounds $\pm \infty$, in $q$-logarithm, either is finite according to $q$.
\begin{align}
    \lim_{x \to 0} \ln_{q}(x) &=
    \begin{cases}
        -\infty & q \geq 1
        \\
        - \frac{1}{1 - q} & q < 1
    \end{cases}
    \label{eq:qlog_lbnd} \\
    \lim_{x \to \infty} \ln_{q}(x) &=
    \begin{cases}
        \frac{1}{q - 1} & q > 1
        \\
        \infty & q \leq 1
    \end{cases}
    \label{eq:qlog_ubnd}
\end{align}
In addition, the following inequality holds for $q_1 < q_2$.
\begin{align}
    \ln_{q_1}(x) \geq \ln_{q_2}(x)
    \label{eq:qlog_ineq}
\end{align}
The equality is satisfied only when $x = 1$.

In the derivation of q-VAE, several important tricks are described below.
First, in $q$-logarithm, pseudo-additivity is established instead of additivity.
\begin{align}
    \ln_{q}(x_1 x_2) &= \ln_{q}(x_1) + \ln_{q}(x_2) + (1 - q) \ln_{q}(x_1) \ln_{q}(x_2)
    \nonumber \\
    &= x_2^{1-q} \ln_{q}(x_1) + \ln_{q}(x_2)
    \label{eq:qlog_padd} \\
    &= \ln_{q}(x_1) + x_1^{1-q} \ln_{q}(x_2)
    \nonumber
\end{align}
For the reciprocal, the following formula holds.
\begin{align}
    \ln_{q}(x^{-1}) = - x^{q-1} \ln_{q}(x)
    \label{eq:qlog_reci}
\end{align}
Finally, the $q$-deformed Kullback-Leibler (KL) divergence (or, Tsallis divergence) is given as follows:
\begin{align}
    \mathrm{KL}_{q}(p_1 \| p_2) = - \int p_1(x) \ln_{q} \frac{p_2(x)}{p_1(x)} dx
    \label{eq:tsallis_div}
\end{align}
Note that some probability distribution models, such as exponential distribution families, have closed-form solutions even for Tsallis divergence~\cite{gil2013renyi}.

\subsection{Original VAE and q-VAE}

As a comparison to q-VAE, we first derive the original VAE~\cite{kingma2014auto}.
For VAE, a dataset $\mathcal{D}_{\mathcal{X}} = \{x_{i}\}_{i=1}^{M}$, where $x \in \mathcal{X}$ is the data observed by sensors and $M$ of them are collected.
Here, $\mathcal{D}_{\mathcal{X}}$ is distinguished from the dataset for the world model $\mathcal{D}_{\mathcal{W}}$ by definition.
However, for practical use, the dataset $\mathcal{D} = \{(x_{i}, a_{i}, x^{\prime}_{i}, r_{i})\}_{i=1}^{N}$ can be reused for both of them by
extracting $\mathcal{D}_{\mathcal{X}} \subset \mathcal{D}$ from it to train VAE;
and converting it into $\mathcal{D}_{\mathcal{W}}$ by mapping $x_{i}, x^{\prime}_{i} \to s_{i}, s^{\prime}_{i}$ with VAE.

Anyway, for $\mathcal{D}_{\mathcal{X}}$, in order to obtain a generative distribution $p(x)$, the problem of maximizing its log-likelihood is considered.
In variational inference, $x$ is supposed to be generated stochastically depending on the corresponding latent variable $z \in \mathcal{Z} \subset \mathbb{R}^{|\mathcal{Z}|}$ (in general, $|\mathcal{Z}| < |\mathcal{X}|$).
In that case, $p(x)$ can be represented to be $p(x) = \int p(x \mid z; \phi) p(z) dz$ with a pre-designed prior distribution $p(z)$ and a decoder $p(x \mid z; \phi)$ with the set of parameters $\phi$.
From this relation, the variational lower bound $- \mathcal{L}(\phi; \mathcal{D}_{\mathcal{X}})$ is derived as follows:
\begin{align}
    \mathbb{E}_{\mathcal{D}_{\mathcal{X}}}[\ln p(x_{i})]
    &= \mathbb{E}_{\mathcal{D}_{\mathcal{X}}} \left[ \ln \int p(x_{i} \mid z; \phi) p(z) dz \right]
    \nonumber \\
    &= \mathbb{E}_{\mathcal{D}_{\mathcal{X}}} \left[ \ln \int p(z \mid x_{i}; \phi) \frac{p(x_{i} \mid z; \phi) p(z)}{p(z \mid x_{i}; \phi)} dz \right]
    \nonumber \\
    &\geq \mathbb{E}_{\mathcal{D}_{\mathcal{X}}, z_{i} \sim p(z \mid x_{i}; \phi)} \left[ \ln p(x_{i} \mid z_{i}; \phi) \right]
    - \mathbb{E}_{\mathcal{D}_{\mathcal{X}}} \left[ \mathrm{KL}(p(z \mid x_{i}; \phi) \| p(z)) \right]
    \nonumber \\
    & = - \mathcal{L}(\phi; \mathcal{D}_{\mathcal{X}})
    \label{eq:vae}
\end{align}
where $p(z \mid x_{i}; \phi)$ denotes the variational posterior distribution (or encoder).
Note that $q(\cdot)$ is generally used instead of $p(\cdot)$ to denote the variational distribution, but since $q$ appears in Tsallis statistics, it is unified with $p(\cdot)$ to avoid confusion.
The inequality in the above derivation is given by Jensen's inequality using the fact that the natural logarithm is a concave function.
In order to minimize $\mathcal{L}(\phi; \mathcal{D}_{\mathcal{X}})$, the computational graph for $\phi$ is constructed using the reparameterization trick~\cite{kingma2014auto}, etc., and one of the stochastic gradient descent methods~\cite{kingma2014adam,ilboudo2020robust} is used to optimize $\phi$.
Furthermore, by considering the minimization problem of $\mathcal{L}(\phi; \mathcal{D}_{\mathcal{X}})$ as a constrained optimization problem with KL divergence, $\beta$-VAE~\cite{higgins2017beta}, which multiplies KL divergence by a weight $\beta > 0$, is derived via Lagrange's method of undetermined multipliers.

For convenience, the first term is called the reconstruction term to increase the accuracy of reconstructing the observed data from the encoded latent variable, and the second term is called the regularization term that attempts to match the encoder to the prior.
The regularization term shapes the latent space according to the prior, and the design of the prior promotes disentangled representation (i.e. independence and sparsification).
For implementation, from the many reasons (e.g. the closed-form solution of KL divergence can be obtained, the reparameterization trick is well established, and the computational cost is small), $p(z)$ is frequently given by the standard normal distribution $\mathcal{N}(0, I)$, and $p(z \mid x_{i}; \phi)$ is accordingly modeled by a diagonal normal distribution.
Note that the model of $p(x \mid z; \phi)$ depends on $x$:
for real data such as robot coordinates, a diagonal normal distribution (with fixed variance in some cases) or other real-space distribution like student-t distribution~\cite{takahashi2018student} is used;
and for image data (normalized to $[0, 1]$ for each pixel), Bernoulli distribution (recently, continuous Bernoulli distribution~\cite{loaiza2019continuous}) is adopted.

Finally, we introduce the original q-VAE~\cite{kobayashi2020q}, which uses the same variables and probability distributions, but replaces the starting point for the maximization problem with the $q$-log likelihood.
By restricting $q > 0$ to make $q$-logarithm concave, the variational lower bound for this problem can be derived as in the usual VAE (note the pseudo-additivity).
\begin{align}
    \mathbb{E}_{\mathcal{D}_{\mathcal{X}}}[\ln_{q} p(x_{i})]
    &= \mathbb{E}_{\mathcal{D}_{\mathcal{X}}} \left[ \ln_{q} \int p(z \mid x_{i}; \phi) \frac{p(x_{i} \mid z; \phi) p(z)}{p(z \mid x_{i}; \phi)} dz \right]
    \nonumber \\
    &\geq \mathbb{E}_{\mathcal{D}_{\mathcal{X}}, z_{i} \sim p(z \mid x_{i}; \phi)} \left[ \rho(x_{i}, z_{i})^{1-q} \ln_{q} p(x_{i} \mid z_{i}; \phi) \right]
    - \mathbb{E}_{\mathcal{D}_{\mathcal{X}}} \left[ \mathrm{KL}_{q}(p(z \mid x_{i}; \phi) \| p(z)) \right]
    \nonumber \\
    & = - \mathcal{L}_{q}(\phi; \mathcal{D}_{\mathcal{X}})
    \label{eq:qvae}
\end{align}
where $\rho(x, z) = p(z) / p(z \mid x; \phi) = (1 - q) \ln_{q} p(z) / p(z \mid x; \phi) + 1 > 0$.
Under the condition on $q < 1$, when $\rho$ is small (i.e. $p(z) < p(z \mid x; \phi)$), the influence of the reconstruction term is suppressed and the regularization term dominates, thus promoting $p(z \mid x; \phi) \to p(z)$.
Otherwise, the reconstruction term becomes dominant and $p(z \mid x; \phi)$ tries to extract the information needed for the reconstruction.
In the original paper, this behavior is regarded as an adaptive $\beta$ in $\beta$-VAE, automatically adjusting the trade-off between the disentangled representation promoted by large $\beta$ and the reconstruction accuracy impaired by it.
Indeed, experimental results in that paper showed that the reconstruction accuracy can be retained while increasing the independence among latent variables compared to $\beta$-VAE.
Note that, with $q = 1$, the above problem reverts to the standard VAE.

\section{Modified q-VAE}

\subsection{Stability issues}

In the original q-VAE, numerical stability issues were found, and two ad-hoc cheap tricks were made to address these issues.
The first is the removal of the computational graph leading to $\rho$, making $\rho$ a merely adaptive coefficient.
In this way, $\rho$ can be regarded as a part of $\beta$ in the original version, but $\phi$ should naturally be updated by the computational graph to $\rho$.
This trick may cause large biases in the behavior during training and the obtained latent space.

The other is the limitation of the decoder model.
In many cases, the observed data handled by VAE are of very high dimension, and the following another representation of $q$-logarithm for them reveals that it tends to have relatively large values in the exponential function, causing numerical divergence.
\begin{align}
    \frac{p(x)^{1-q} - 1}{1 - q} &= \frac{\exp\{(1 - q) \ln p(x)\} - 1}{1 - q}
    \nonumber \\
    &= \frac{\exp\{(1 - q) \sum_{k=1}^{|\mathcal{X}|} \ln p(x_{k})\} - 1}{1 - q}
    \label{eq:qlog2}
\end{align}
That is, if $p(x)$ is given as probability density function (i.e. $x$ is in continuous space), $p(x)$ can be over one, resulting in the positive log-likelihood.
Even if the log-likelihood for each dimension is slightly positive, the value accumulated tens of thousands of them would easily diverge the above exponential function numerically.
To avoid this issue, the original q-VAE limits the decoder model that cancel $q$-logarithm, such as $q$-Gaussian distribution~\cite{suyari2005law}.
Of course, it is not a good idea to do so, because various literatures have reported performance gains by utilizing different decoder models~\cite{takahashi2018student,loaiza2019continuous}, as mentioned above.

For these two open issues, this paper deepens the analysis of the original q-VAE and decomposes it into a new surrogated variational lower bound.
In addition, as a part of the flexibility of the decoder model, we also derive a formulation that takes into account the case of mixed observations that should be represented by different models.
Note that we take care of making the modified q-VAE a general form by guaranteeing that it reverts to the standard VAE when $q = 1$ (and other hyperparameters are appropriately given).

\subsubsection{Alternative to removal of computational graph}

First, Tsallis divergence is decomposed as follows, making full use of its definition in eq.~\eqref{eq:tsallis_div}, pseudo-additivity in eq.~\eqref{eq:qlog_padd}, and the formula for the reciprocal in eq.~\eqref{eq:qlog_reci}.
\begin{align}
    \mathrm{KL}_{q}(p_1 \| p_2) &= - \int p_1(x) \ln_{q} \frac{p_2(x)}{p_1(x)} dx
    \nonumber \\
    &= - \mathbb{E}_{p_1}[p_1(x)^{q-1} \ln_{q} p_2(x) + \ln_{q} p_1(x)^{-1}]
    \nonumber \\
    &= - \mathbb{E}_{p_1}[p_1(x)^{q-1} \ln_{q} p_2(x) - p_1(x)^{q-1} \ln_{q} p_1(x)]
    \nonumber \\
    &= - \mathbb{E}_{p_1}[p_1(x)^{q-1} \{\ln_{q} p_2(x) - \ln_{q} p_1(x)\}]
\end{align}
By applying this to eq.~\eqref{eq:qvae} and decomposing $\rho$ as $p(z)/p(z \mid x; \phi)$, we see that $p(z \mid x; \phi)^{q-1}$ is multiplied to every term.
\begin{align}
    \eqref{eq:qvae} &= \mathbb{E}_{\mathcal{D}_{\mathcal{X}}, z_{i} \sim p(z \mid x_{i}; \phi)} \left[
    \rho(x_{i}, z_{i})^{1-q} \ln_{q} p(x_{i} \mid z_{i}; \phi)
    + p(z_{i} \mid x_{i}; \phi)^{q-1} \left \{ \ln_{q} p(z_{i}) - \ln_{q} p(z_{i} \mid x_{i}; \phi) \right\} \right]
    \nonumber \\
    &= \mathbb{E}_{\mathcal{D}_{\mathcal{X}}, z_{i} \sim p(z \mid x_{i}; \phi)} \left[
    p(z_{i} \mid x_{i}; \phi)^{q-1} \left \{ p(z_{i})^{1-q} \ln_{q} p(x_{i} \mid z_{i}; \phi) + \ln_{q} p(z_{i}) - \ln_{q} p(z_{i} \mid x_{i}; \phi) \right\} \right]
    \label{eq:qvae2}
\end{align}
Here, with the fact $p(z_{i} \mid x_{i}; \phi) > 0$ and $p(z_{i} \mid x_{i}; \phi)^{q-1} \simeq 1$ when $q \simeq 1$, $p(z_{i} \mid x_{i}; \phi)^{q-1}$ can be ignored as a slightly biased but consistent surrogated problem.
\begin{align}
    \eqref{eq:qvae2} \propto \mathbb{E}_{\mathcal{D}_{\mathcal{X}}, z_{i} \sim p(z \mid x_{i}; \phi)} \left[
    p(z_{i})^{1-q} \ln_{q} p(x_{i} \mid z_{i}; \phi) + \ln_{q} p(z_{i}) - \ln_{q} p(z_{i} \mid x_{i}; \phi) \right]
    \label{eq:qvae3}
\end{align}
In this case, the first term can be regarded as the reconstruction term, the second as the regularization term that brings the encoder closer to the prior, and the third as an entropy term that maximizes the entropy of the encoder.
Although the surrogated problem induces only a small bias, the elimination of $p(z_{i} \mid x_{i}; \phi)^{q-1}$ simplifies the gradient considerably, making it numerically much more stable.
Note that the reconstruction term is not yet sufficiently stable numerically at this stage, since the direction to be updated is not unique unless under certain conditions, as described later.

\subsubsection{Alternative to limitation of decoder model}

Based on eq.~\eqref{eq:qvae3}, the decoder model is first decomposed for representing several types of observations.
To this end, $x \in \mathcal{X}$ is classified into $C \in \mathbb{N}$ classes, i.e. $x = [x_{1}, x_{2}, \ldots, x_{C}]$ with $x_{c} \in \mathcal{X}_{c}$ ($c = 1, 2, \ldots, C$).
Note that, at this stage, each class is unordered.
Suppose that each class is independent, the decoder can be decomposed as follows:
\begin{align}
    p(x \mid z; \phi) = \prod_{c=1}^{C} p(x_{c} \mid z; \phi)
\end{align}
where $p(x_{c} \mid z; \phi)$ is modeled by an appropriate distribution, such as a diagonal normal distribution and a continuous Bernoulli distribution.
Although the production of likelihoods can be converted into the sum of them through natural logarithm, $q$-logarithm requires the pseudo-additivity defined in eq.~\eqref{eq:qlog_padd}.
By applying the pseudo-additivity iteratively, $q$-logarithm of the decomposed decoders is derived as follows:
\begin{align}
    \ln_{q} p(x \mid z; \phi) = \sum_{c=1}^{C} p(x_{< c} \mid z; \phi)^{1-q} \ln_{q} p(x_{c} \mid z; \phi)
\end{align}
where
\begin{align}
    p(x_{< c} \mid z; \phi) =
    \begin{cases}
        1 & c = 1
        \\
        \prod_{j=1}^{c-1} p(x_{j} \mid z; \phi) & c > 1
    \end{cases}
\end{align}

To avoid numerical divergence of the decomposed decoders, we pay attention to $1-q$, which is in the exponential function of eq.~\eqref{eq:qlog2} as a coefficient.
That is, the larger $q$ is, the smaller the scale of values in the exponential function becomes.
Therefore, the condition for no numerical divergence can be found by increasing $q$.
In addition, as introduced in eq.~\eqref{eq:qlog_ineq}, the $q$-logarithm becomes smaller for larger $q$.
From these facts, the following lower bound is gained by introducing $q_{c}$ ($q \leq q_{1} \leq q_{2} \leq \ldots \leq q_{C} \leq 1$).
\begin{align}
    \ln_{q} p(x \mid z; \phi) &\geq \sum_{c=1}^{C} p(x_{< c} \mid z; \phi)^{1-q_{1}} \ln_{q_{1}} p(x_{c} \mid z; \phi)
    \nonumber \\
    &= \ln_{q_{1}} p(x_{1} \mid z; \phi) + p(x_{1} \mid z; \phi)^{1-q_{1}} \sum_{c=2}^{C} p(x_{< c} \mid z; \phi)^{1-q_{1}} \ln_{q_{1}} p(x_{c} \mid z; \phi)
    \nonumber \\
    &\geq \ln_{q_{1}} p(x_{1} \mid z; \phi) + p(x_{1} \mid z; \phi)^{1-q_{1}} \sum_{c=2}^{C} p(x_{< c} \mid z; \phi)^{1-q_{2}} \ln_{q_{2}} p(x_{c} \mid z; \phi)
    \nonumber \\
    &\geq \sum_{c=1}^{C} p(x_{< c} \mid z; \phi)^{1-q_{< c}} \ln_{q_{c}} p(x_{c} \mid z; \phi)
\end{align}
where
\begin{align}
    p(x_{< c} \mid z; \phi)^{1-q_{< c}} =
    \begin{cases}
        1 & c = 1
        \\
        \prod_{j=1}^{c-1} p(x_{j} \mid z; \phi)^{1-q_{j}} & c > 1
    \end{cases}
\end{align}
Note that although $x_{c}$ was assumed to be unordered, its order can be determined by finding each of the non-divergent $q_{c}$ and arranging them in ascending order.
As the continuous space with larger $|\mathcal{X}_c|$ tends to diverge more easily, this fact can be a guide for roughly adjusting $q_{c}$.

By substituting this lower bound into eq.~\eqref{eq:qvae3}, a modified q-VAE, which can be numerically stable under the appropriate conditions, is obtained.
Here, for convenience, all terms are weighted respectively as in $\beta$-VAE~\cite{higgins2017beta} and its variants like~\cite{mathieu2019disentangling}.
Specifically, the modified q-VAE aims to find $\phi$ that maximizes the following equation (i.e. minimizes $\mathcal{\tilde{L}}_{q}(\phi; \mathcal{D}_{\mathcal{X}})$) weighted by $\zeta_{c} > 0$, $\beta > 0$, and $\gamma > 0$.
\begin{align}
    \eqref{eq:qvae3} &\geq \mathbb{E}_{\mathcal{D}_{\mathcal{X}}, z_{i} \sim p(z \mid x_{i}; \phi)} \Biggl[
    p(z_{i})^{1-q} \sum_{c=1}^{C} p(x_{< c} \mid z; \phi)^{1-q_{< c}} \ln_{q_{c}} p(x_{c} \mid z; \phi)
    \nonumber \\
    &\qquad \qquad \qquad \qquad + \ln_{q} p(z_{i}) - \ln_{q} p(z_{i} \mid x_{i}; \phi) \Biggr]
    \nonumber \\
    &\propto \mathbb{E}_{\mathcal{D}_{\mathcal{X}}, z_{i} \sim p(z \mid x_{i}; \phi)} \Biggl[
    p(z_{i})^{1-q} \sum_{c=1}^{C} \zeta_{c} p(x_{< c} \mid z; \phi)^{1-q_{< c}} \ln_{q_{c}} p(x_{c} \mid z; \phi)
    \nonumber \\
    &\qquad \qquad \qquad \qquad + \beta \ln_{q} p(z_{i}) - \gamma \ln_{q} p(z_{i} \mid x_{i}; \phi) \Biggr]
    \nonumber \\
    &= - \mathcal{\tilde{L}}_{q}(\phi; \mathcal{D}_{\mathcal{X}})
    \label{eq:mqvae}
\end{align}

\subsection{Analysis for sparsification}

In the original q-VAE, the computational graph of $\rho$ is removed to stabilize the computation, and it becomes a merely adaptive coefficient.
Thus, overall, the original q-VAE solves the multi-objective optimization problem of reconstruction and regularization terms by scalarizing them with an (adaptive) linear weighted \textit{sum}.
On the other hand, in the modified q-VAE (with $q < 1$), the likelihood of the encoder, which was the denominator of $\rho$, is eliminated, but the regularization by the prior distribution in the numerator is retained, preserving the computational graph.
This means that the multi-objective optimization problem with the reconstruction and regularization terms is scalarized and solved in the form of a \textit{product}.
This product is important: both terms must be satisfied simultaneously to obtain a high value.

However, in the maximization problem involving such a product, each term must always be non-negative.
That is, if the sign of one term is reversed to negative, the other term has a negative coefficient, converting its maximization problem into a minimization problem.
Of course, this switching is also a factor that makes learning unstable.
Since the regularization term $p(z_{i})^{1-q}$ is non-negative in definition, we need to reveal the conditions for the non-negative reconstruction term.

Now, we first consider the case with $C = 1$ for simplicity.
The reconstruction term $\ln_{q_{1}} p(x_{1} \mid z; \phi)$ becomes negative with $p(x_{1} \mid z; \phi) < 1$ and non-negative with $p(x_{1} \mid z; \phi) \geq 1$.
Since sign reversal may occur depending on the performance of the decoder, we substitute the definition of $q$-logarithm when $q \neq 1$ (see eq.~\eqref{eq:qlog}) and reform eq.~\eqref{eq:mqvae} as follows:
\begin{align}
    &\mathbb{E}_{\mathcal{D}_{\mathcal{X}}, z_{i} \sim p(z \mid x_{i}; \phi)} \Biggl[
    p(z_{i})^{1-q} \zeta_{1} \ln_{q_{1}} p(x_{1} \mid z; \phi) + \beta \ln_{q} p(z_{i}) - \gamma \ln_{q} p(z_{i} \mid x_{i}; \phi) \Biggr]
    \nonumber \\
    =& \mathbb{E}_{\mathcal{D}_{\mathcal{X}}, z_{i} \sim p(z \mid x_{i}; \phi)} \Biggl[
    p(z_{i})^{1-q} \left \{ \zeta_{1} \ln_{q_{1}} p(x_{1} \mid z; \phi) + \frac{\beta}{1 - q} \right \} - \frac{\beta}{1 - q} - \gamma \ln_{q} p(z_{i} \mid x_{i}; \phi) \Biggr]
    \label{eq:mqvae_anl1}
\end{align}
Using this form and the fact that $q$-logarithm with $q < 1$ has a finite lower bound as shown in eq.~\eqref{eq:qlog_lbnd}, the condition for the values in the brace $\{\cdot\}$ to always be non-negative is revealed.
\begin{align}
     - \frac{\zeta_{1}}{1 - q_{1}} + \frac{\beta}{1 - q} \geq 0
    \nonumber \\
    \therefore \beta \geq \frac{1 - q}{1 - q_{1}} \zeta_{1}
\end{align}

Similarly, the conditions required when $C > 1$ are also considered.
The terms inside the brace $\{\cdot\}$ of eq.~\eqref{eq:mqvae_anl1} is derived as follows:
\begin{align}
    &\zeta_{C} p(x_{< C} \mid z; \phi)^{1-q_{< C}} \frac{p(x_{C} \mid z; \phi)^{1-q_{C}}}{1 - q_{C}}
    \nonumber \\
    +& \sum_{c=1}^{C} p(x_{< c} \mid z; \phi)^{1-q_{< c}} \left ( \frac{\zeta_{c-1}}{1 - q_{c-1}} - \frac{\zeta_{c}}{1 - q_{c}} \right )
\end{align}
where $q_{0} = q$ and $\zeta_{0} = \beta$ for simplifying the description.
The first term is always non-negative, so the required conditions are gained from the second term.
\begin{align}
    \frac{\zeta_{c-1}}{1 - q_{c-1}} - \frac{\zeta_{c}}{1 - q_{c}} \geq 0, \ c = 1, 2, \ldots, C
    \nonumber \\
    \therefore \beta \geq \frac{1 - q}{1 - q_{1}} \zeta_{1} \geq \frac{1 - q}{1 - q_{2}} \zeta_{2} \geq \ldots \geq \frac{1 - q}{1 - q_{C}} \zeta_{C}
    \label{eq:condition}
\end{align}

When the above conditions are satisfied, the improvement in the reconstruction accuracy of the observed data (i.e. compression of important information into the latent space) and the regularization of the encoder (i.e. organization of the latent space) should be promoted simultaneously.
It is also important that this regularization is given in terms of likelihood rather than log-likelihood to the prior.
That is, even if the prior is assumed to be a diagonal distribution such as $p(z) = \mathcal{N}(0, I)$, the likelihood is given by the product of the likelihoods for the respective dimensions.
As in the discussion above, if the regularization for each dimension is taken as a multi-objective problem, the regularization for all dimensions should be simultaneously established.
This suggests that each dimension of the latent space is sparsified as much as possible to match the prior (basically, centered at the origin), while still extracting sufficient information into the latent space to reconstruct the observed data.
In addition, the regularization status of the other dimensions influences each other, thus preventing duplication of information and encouraging independence.
As a result, the latent variable $z$ on the latent space constructed by the modified q-VAE is expected to coincide with the state $s$, which is the minimal realization for the whole behavior of the system.

\section{Simulation}

\subsection{Task}

\begin{figure}[tb]
    \centering
    \includegraphics[keepaspectratio=true,width=0.96\linewidth]{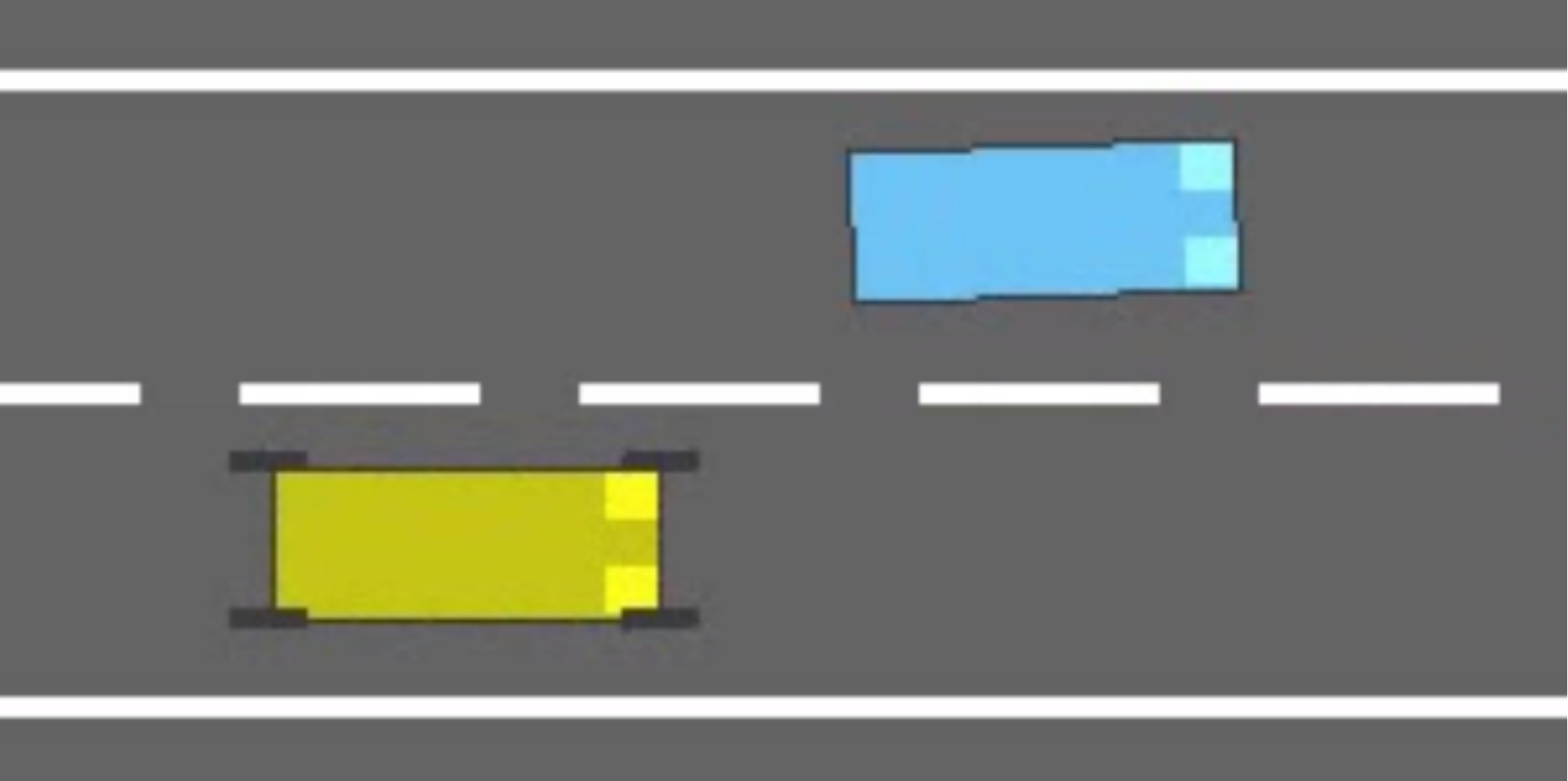}
    \caption{\textit{highway-env}:
        an agent with yellow aims to drive within the lanes while avoiding other blue cars.
    }
    \label{fig:sim_env}
\end{figure}

\begin{figure}[tb]
    \centering
    \includegraphics[keepaspectratio=true,width=0.96\linewidth]{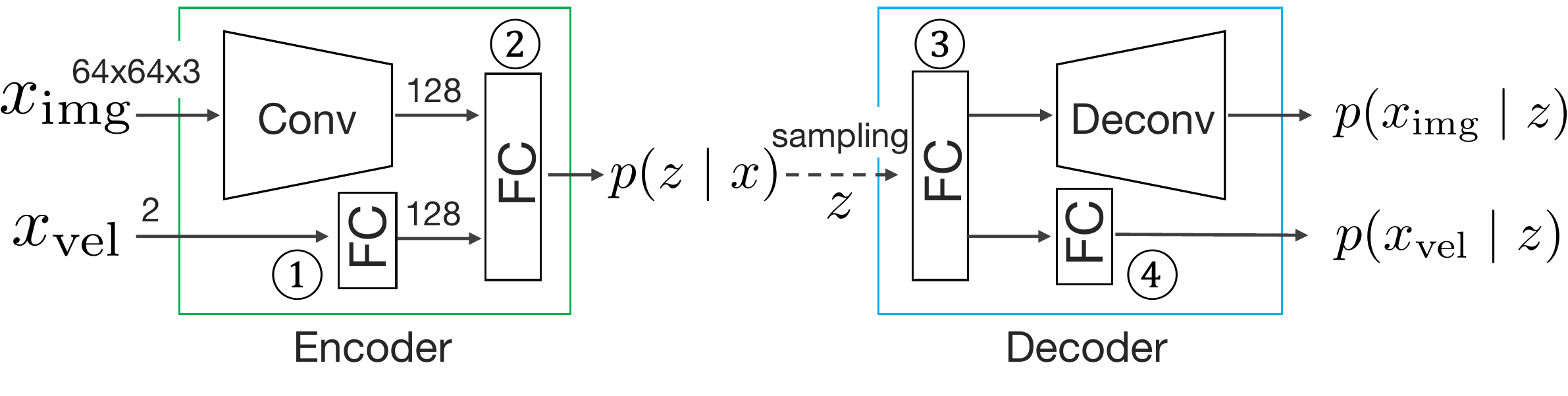}
    \caption{Network architecture of VAE:
        after encoding the image $x_{\mathrm{img}}$ and velocity $x_{\mathrm{vel}}$ separately to some extent, the two are concatenated and further encoded;
        from the obtained posterior distribution $p(z \mid x)$, the two observations are reconstructed separately after sampling the latent variable $z$ (with the reparameterization trick);
        details of the respective modules \textcircled{1}--\textcircled{4} are shown in Table~\ref{tab:net_vae}.
    }
    \label{fig:impl_vae}
\end{figure}

As a statistical validation in the simulation, we use CEM to control \textit{highway-env}~\cite{highway-env}.
Specifically, as shown in Fig.~\ref{fig:sim_env}, this task is to avoid colliding with other blue car(s) and going out of the lane by operating the accelerator and steering wheel (i.e. two-dimensional continuous action space) of a yellow car.
Usually, geometric information between cars can be used for observation, but in this verification, the RGB image of 300$\times$150$\times$3 in Fig.~\ref{fig:sim_env} is resized to 64$\times$64$\times$3 as an observation.
This modification requires to extract the latent state from high-dimensional observation.
In addition, to show that multiple types of sensors can be integrated as described above, the yellow car's velocity $v_{x,y}$ is given as another observation.

The control by CEM is real-time oriented and outputs the currently optimal action in each control period even if the maximum number of iterations has not been finished.
In addition, the network architecture of VAE implemented by PyTorch~\cite{paszke2017automatic} is illustrated in Fig.~\ref{fig:impl_vae}.
These details and the way to collect the dataset are described in Appendix~\ref{app:config}.

\subsection{Results}

\begin{table}[tb]
    \caption{Comparisons for \textit{highway-env}}
    \label{tab:comp_highway}
    \centering
    \begin{tabular}{cccccccc}
        \hline\hline
        Method & $q$ & $q_1$ & $q_2$ & $\zeta_1$ & $\zeta_2$ & $\beta$ & $\gamma$
        \\
        \hline
        VAE & 1 & -- & -- & 50 & 1 & 1 & --
        \\
        $\beta$-VAE & 1 & -- & -- & 50 & 1 & 0.3 & --
        \\
        q-VAE (proposal) & 0.99 & 0.99 & 0.999 & 10 & 1 & 10 & 3
        \\
        \hline\hline
    \end{tabular}
\end{table}

Under the common settings described above, three methods shown in Table~\ref{tab:comp_highway} are compared.
Note that these values were adjusted by trial and error to increase the accuracy of observation reconstruction as much as possible.
In addition, q-VAE is set to satisfy the sparsity condition indicated by eq.~\eqref{eq:condition} while confirming the stability of the numerical computation.

\subsection{Sparse extraction of latent state}

\begin{figure*}[tb]
    \begin{subfigure}[b]{0.48\linewidth}
        \centering
        \includegraphics[keepaspectratio=true,width=\linewidth]{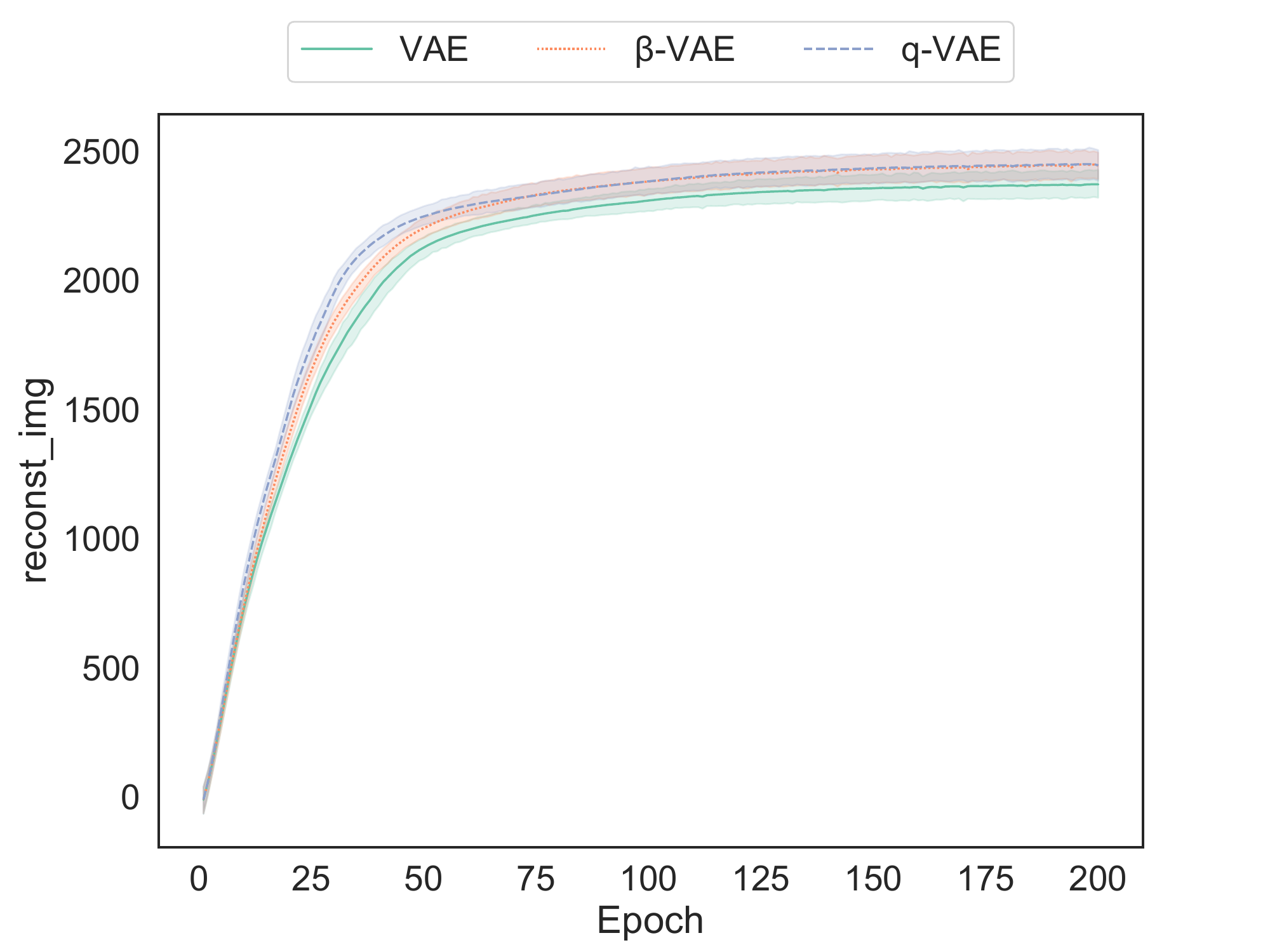}
        \subcaption{Image $x_{\mathrm{img}}$}
        \label{fig:sim_learn_img}
    \end{subfigure}
    \begin{subfigure}[b]{0.48\linewidth}
        \centering
        \includegraphics[keepaspectratio=true,width=\linewidth]{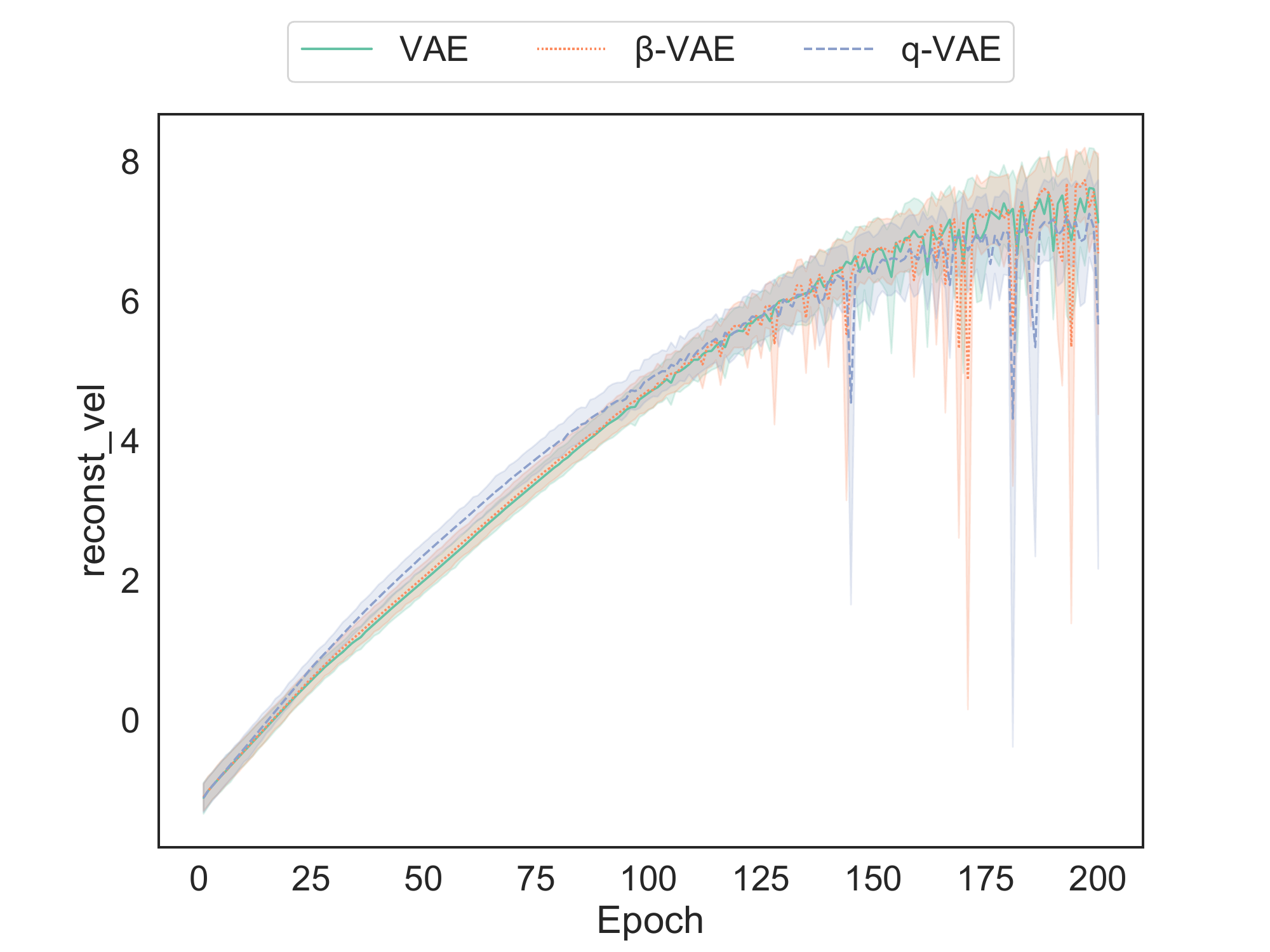}
        \subcaption{Velocity $x_{\mathrm{vel}}$}
        \label{fig:sim_learn_vel}
    \end{subfigure}
    \caption{Learning curves for the reconstruction performance:
        the standard VAE insufficiently reconstructs $x_{\mathrm{img}}$;
        q-VAE improves the reconstruction performance faster than other methods.
    }
    \label{fig:sim_learn}
\end{figure*}

\begin{figure*}[tb]
    \begin{subfigure}[b]{0.48\linewidth}
        \centering
        \includegraphics[keepaspectratio=true,width=\linewidth]{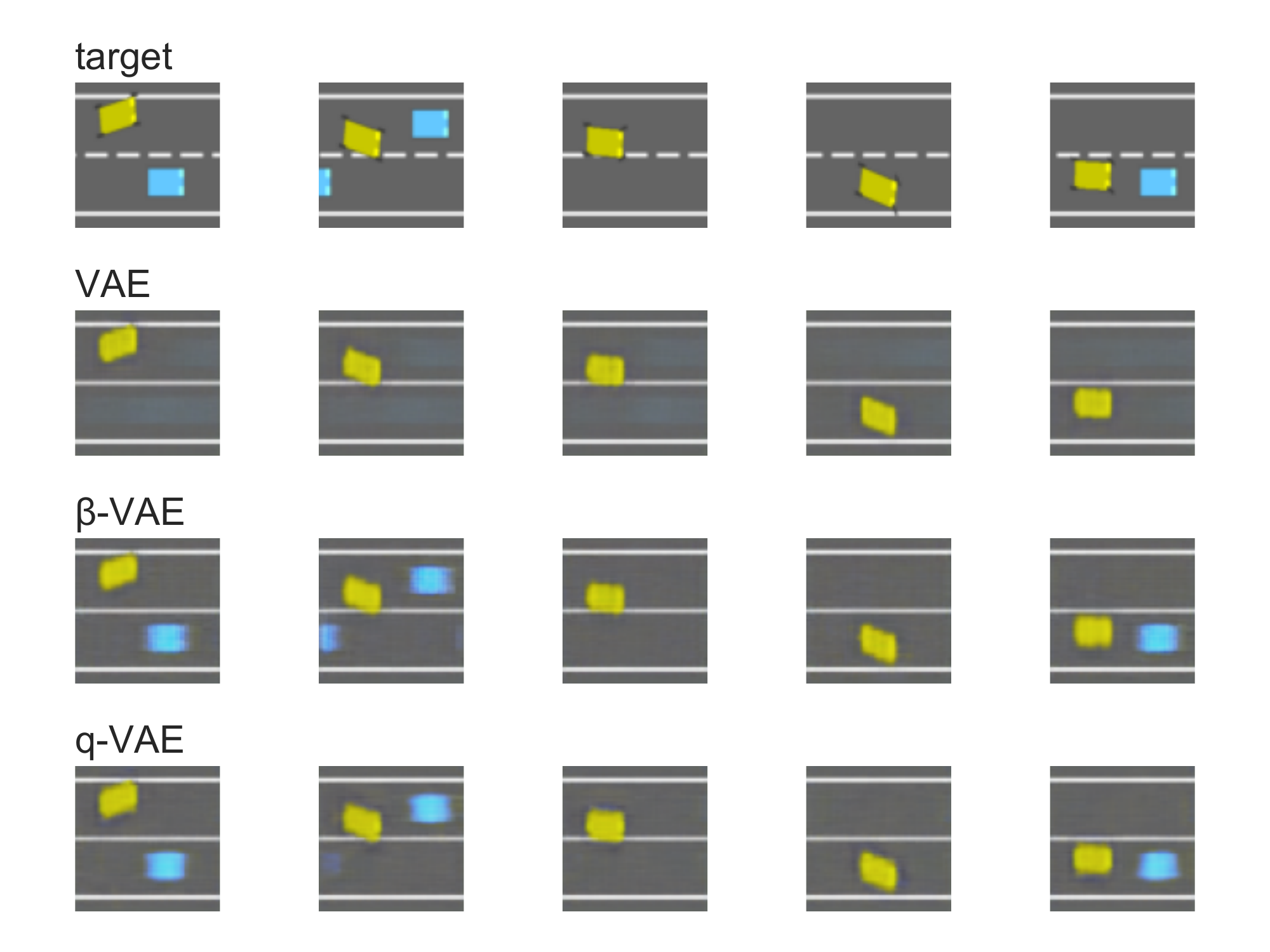}
        \subcaption{Image $x_{\mathrm{img}}$}
        \label{fig:sim_recon_img}
    \end{subfigure}
    \begin{subfigure}[b]{0.48\linewidth}
        \centering
        \includegraphics[keepaspectratio=true,width=\linewidth]{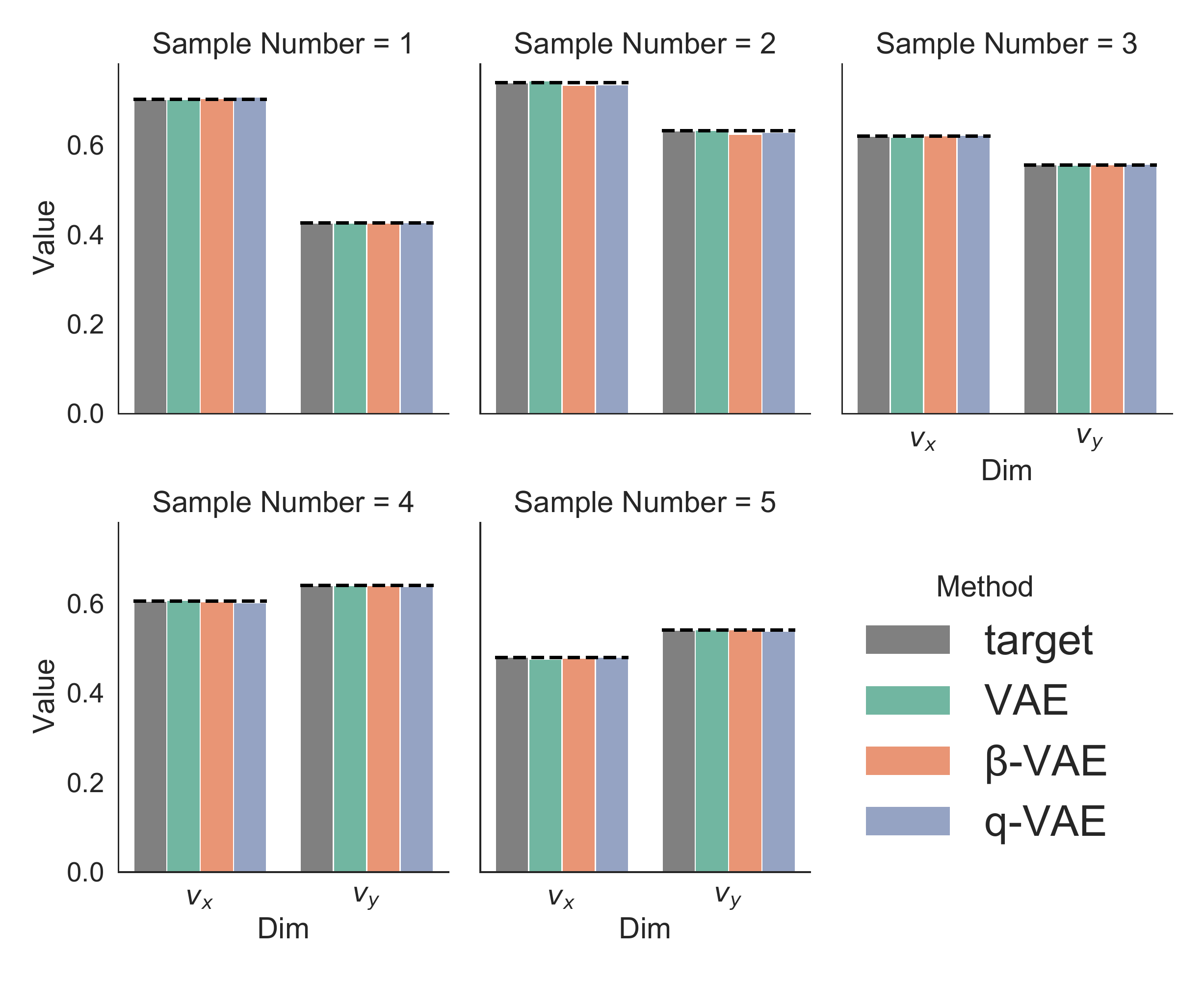}
        \subcaption{Velocity $x_{\mathrm{vel}}$}
        \label{fig:sim_recon_vel}
    \end{subfigure}
    \caption{Examples of the five reconstructed observations:
        the standard VAE failed to reconstruct the blue other cars;
        The remaining two methods were able to reconstruct both observations ($x_{\mathrm{img}}$ and $x_{\mathrm{vel}}$) to the same degree.
    }
    \label{fig:sim_recon}
\end{figure*}

\begin{figure}[tb]
    \centering
    \includegraphics[keepaspectratio=true,width=0.96\linewidth]{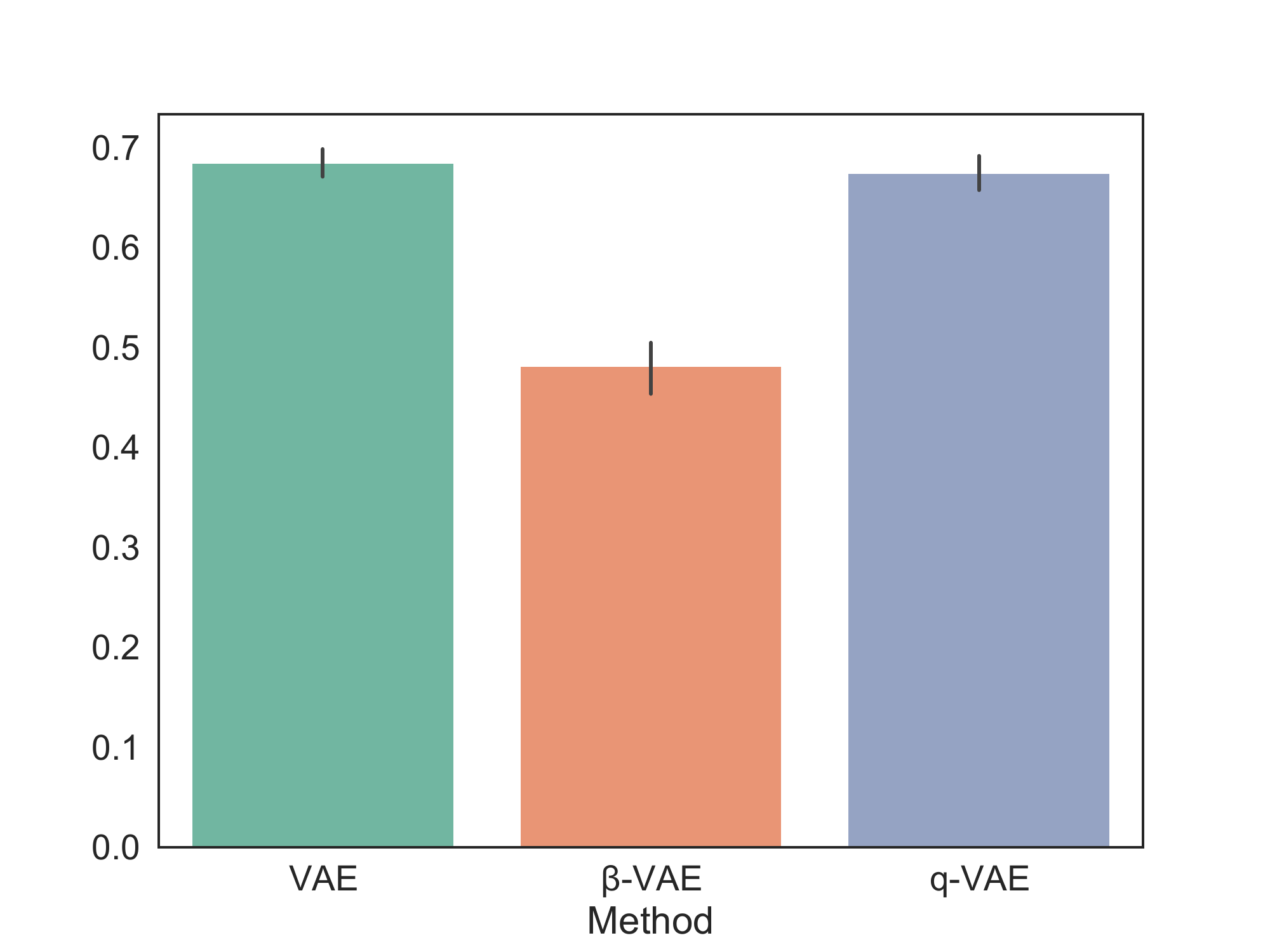}
    \caption{Sparsity of the latent space:
        $\beta$-VAE lost its sparsity, although it improved the reconstruction performance from the standard VAE;
        q-VAE achieved higher sparsity than $\beta$-VAE, while holding the same level of reconstruction;
        the sparsity of the standard VAE was at the expense of reconstruction performance.
    }
    \label{fig:sim_sparse}
\end{figure}

The learning curves for the statistical reconstruction performance (for image and velocity, respectively) with 25 trials are depicted in Fig.~\ref{fig:sim_learn}.
The well-tuned parameters yielded that $\beta$-VAE and q-VAE eventually achieved approximately the same level of reconstruction performance, while the standard VAE (with $\beta = 1$) performed poorly in reconstructing images.
This is probably due to the strong regularization to the prior distribution $p(z)$.
In fact, $\beta$-VAE succeeded in image reconstruction by setting $\beta = 0.3 < 1$.

Another feature of q-VAE is that its learning speed tends to be faster than others.
Although it is difficult to make a clear agreement because the parameter settings differ greatly from others,
we can consider that $\gamma$ over the entropy term of the encoder can be set separately from $\beta$, and $\gamma < \beta$ mitigates the loss of information from the encoder (and the latent space).
In fact, previous studies have pointed out the negative effects of the entropy term found in the decomposition of the regularization term in VAE~\cite{mathieu2019disentangling}, and this is consistent with those reports.

Next, five samples are selected to illustrate the post-learning reconstruction accuracy in Fig.~(\ref{fig:sim_recon}).
As can be seen, while all methods succeeded in reconstructing the velocity with good accuracy, in the image reconstruction, the standard VAE in the second row failed to visualize the other blue car(s).
In other words, it can be said that the encoder of the standard VAE did not properly incorporate the information of the other blue cars into the latent state.
In contrast, $\beta$-VAE and q-VAE properly embedded the important information needed for the reconstruction contained in the observation into the latent space.

Finally, the sparsity of the acquired latent space is evaluated.
As sparsity, we use the following definition in the literature~\cite{hoyer2004non}.
\begin{align}
    \mathrm{sparse}(\mathcal{Z}) &= \frac{1}{N} \sum_{i=1}^N \frac{\sqrt{|\mathcal{Z}|} - \mathrm{ratio}(z_i)}{\sqrt{|\mathcal{Z}|} - 1}
    \\
    \mathrm{ratio}(z_i) &= \frac{\sum_{j=1}^{|\mathcal{Z}|} |z_{i,j}|}{\sqrt{\sum_{j=1}^{|\mathcal{Z}|}z_{i,j}^2}}
    \nonumber
\end{align}
where $z_i = \mathbb{E}_{p(z \mid x_i; \phi)}[z]$ denotes the location of the encoder over $x_i$ in the dataset.
If each component of $z_i$ is with the same magnitude, this definition returns zero; in contrast, if only one component has a non-zero value and others are zero, $\mathrm{sparse}(\mathcal{Z})$ converges to one.

With this definition of the sparsity, we evaluated each method as shown in Fig.~\ref{fig:sim_sparse}.
As a result, only $\beta$-VAE gained low sparsity.
This is due to the fact that $\beta < 1$ is used to improve the reconstruction accuracy.
The standard VAE with $\beta = 1$ achieved the same level of sparsity as the proposed q-VAE, but at the expense of the reconstruction accuracy as mentioned above, extracting the meaningless latent space.
This trend is also consistent with the previous study~\cite{kobayashi2020q}.
From the above, it can be concluded that q-VAE mitigates the trade-off between the reconstruction accuracy and sparsity, and increases both of them sufficiently, namely, it enables to acquire the important information contained in the observation with the smallest dimension size of the latent space (i.e. minimal realization).

\subsection{Control performance}

\begin{figure*}[tb]
    \begin{subfigure}[b]{0.48\linewidth}
        \centering
        \includegraphics[keepaspectratio=true,width=\linewidth]{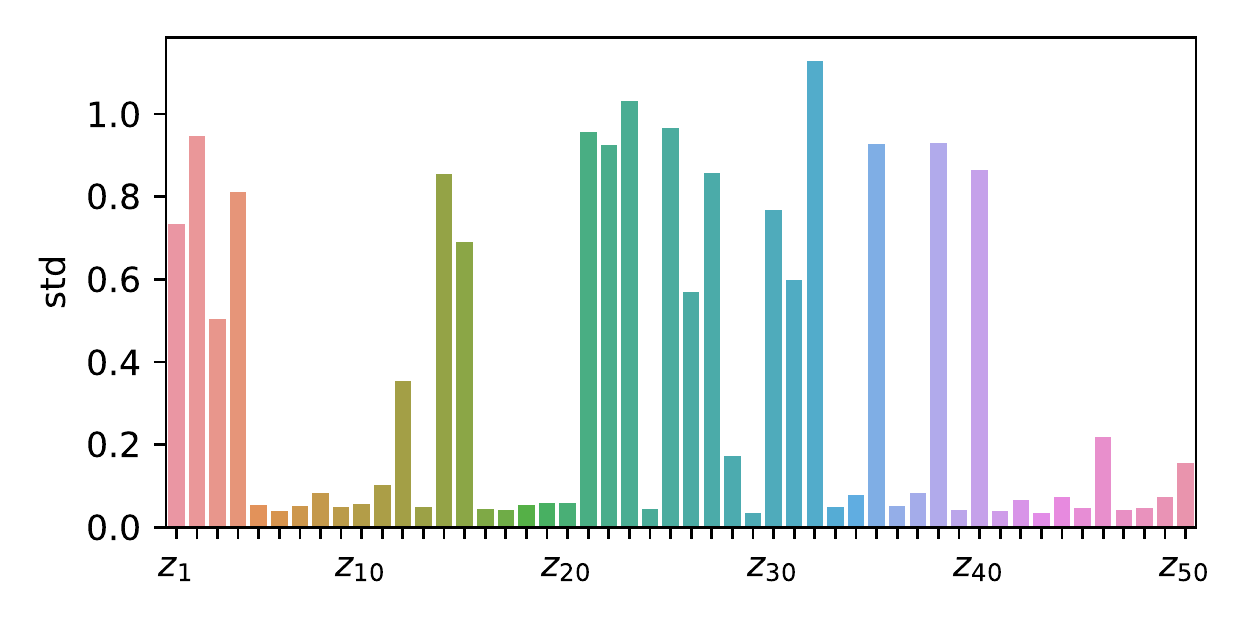}
        \subcaption{$\beta$-VAE}
        \label{fig:sim_latent_bvae}
    \end{subfigure}
    \begin{subfigure}[b]{0.48\linewidth}
        \centering
        \includegraphics[keepaspectratio=true,width=\linewidth]{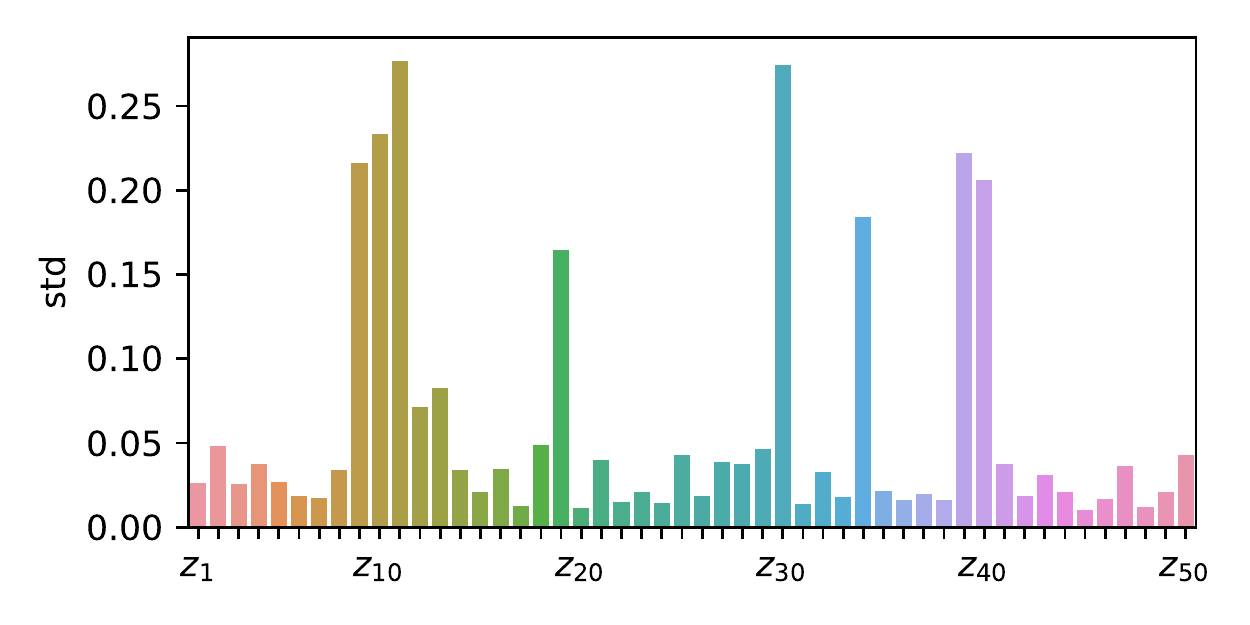}
        \subcaption{q-VAE}
        \label{fig:sim_latent_qvae}
    \end{subfigure}
    \caption{Importance of the latent dimensions:
        (a) $\beta$-VAE remained around half of latent dimensions as importance ones due to low sparsity;
        (b) q-VAE revealed that only eight dimensions (with over 0.15 standard deviation) are important for this task by collapsing other dimensions through sparsification.
    }
    \label{fig:sim_latent}
\end{figure*}

\begin{figure*}[tb]
    \begin{subfigure}[b]{0.48\linewidth}
        \centering
        \includegraphics[keepaspectratio=true,width=\linewidth]{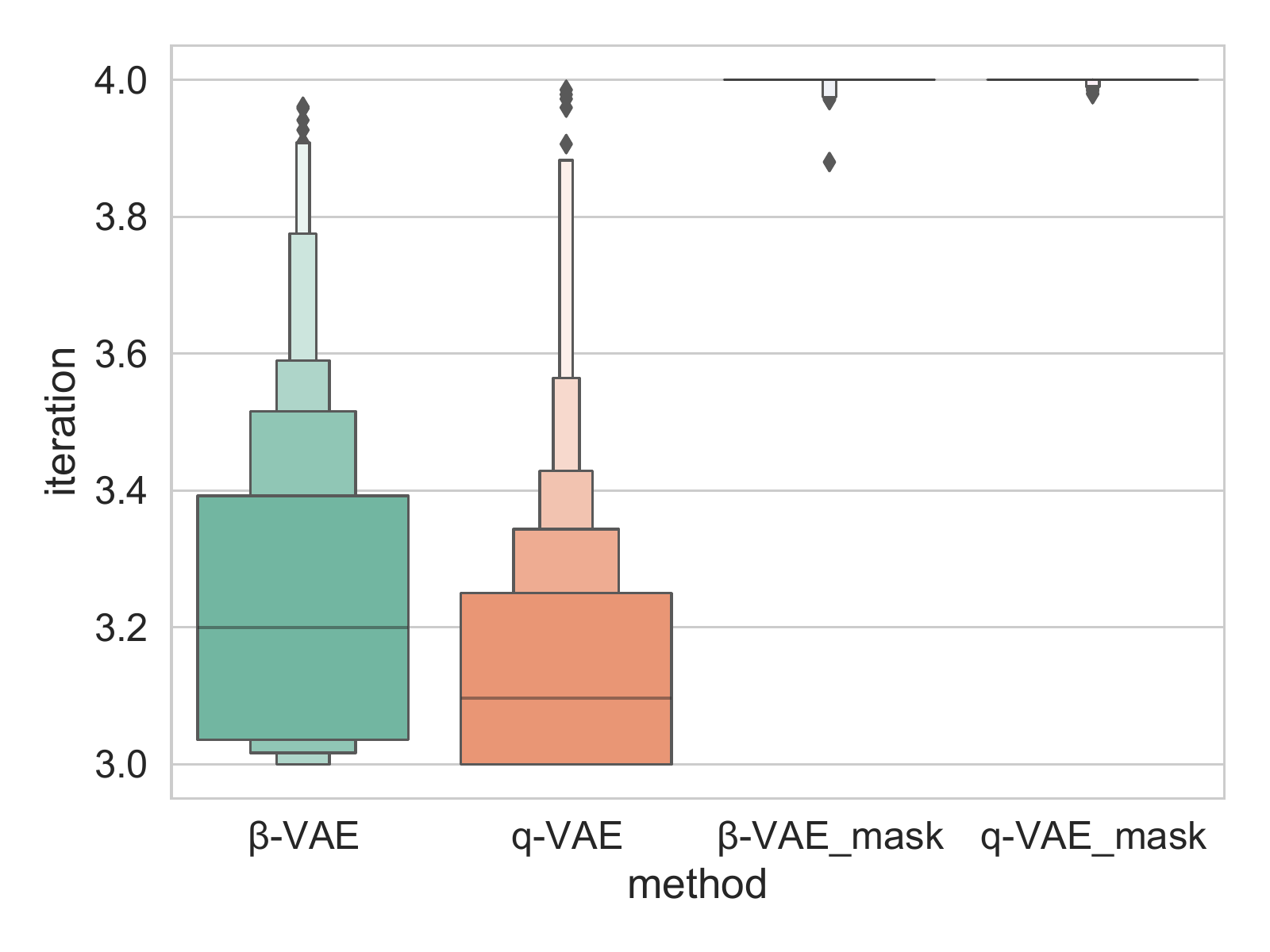}
        \subcaption{\#Iteration}
        \label{fig:sim_ctrl_iter}
    \end{subfigure}
    \begin{subfigure}[b]{0.48\linewidth}
        \centering
        \includegraphics[keepaspectratio=true,width=\linewidth]{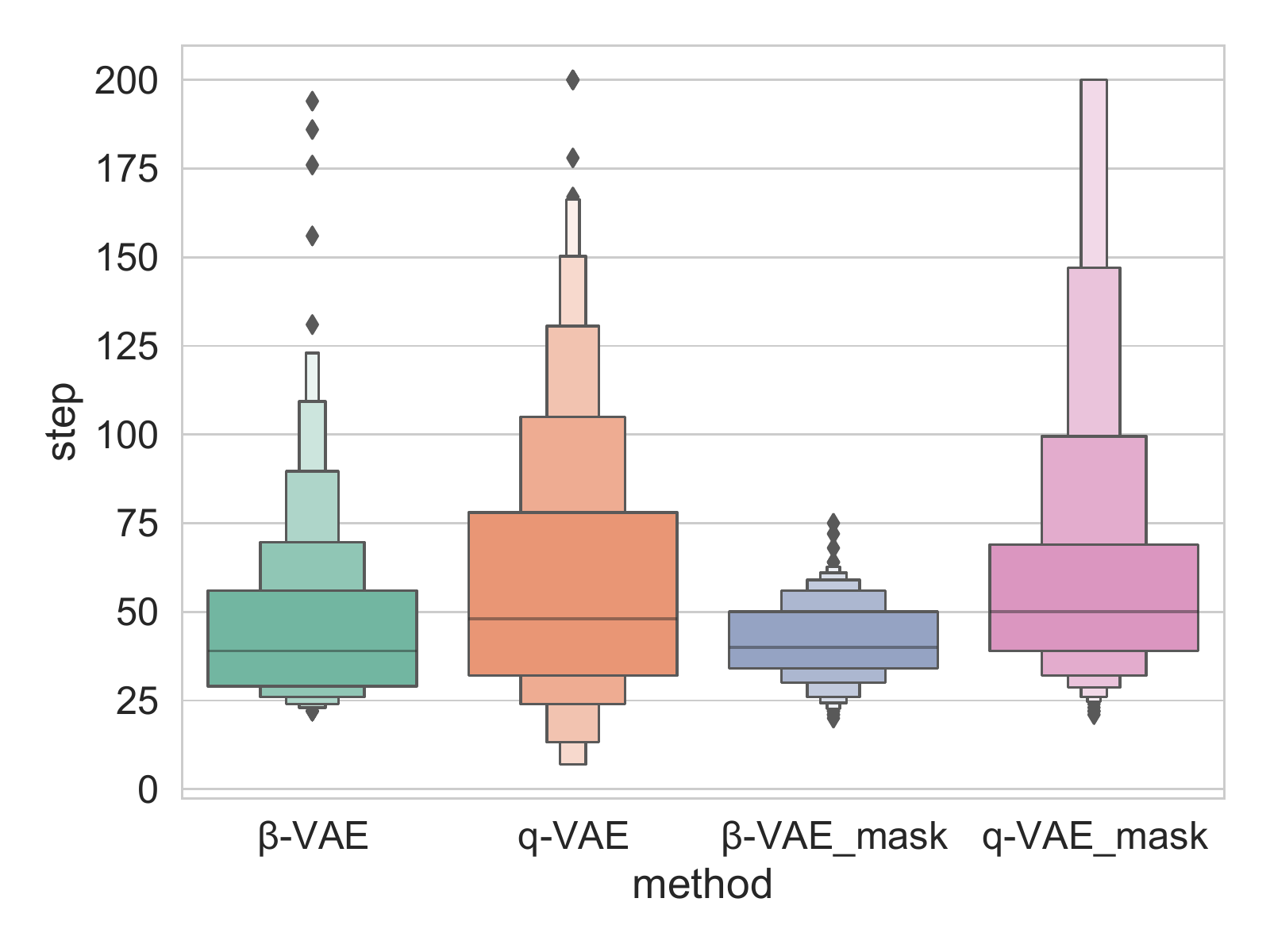}
        \subcaption{\#Step}
        \label{fig:sim_ctrl_step}
    \end{subfigure}
    \caption{Control performance:
        (a) with masking, MPC could iterates its optimization process four times stably;
        (b) the masked q-VAE achieved the better performance than others.
    }
    \label{fig:sim_ctrl}
\end{figure*}

\begin{table}[tb]
    \caption{Network size for \textit{highway-env}}
    \label{tab:size}
    \centering
    \begin{tabular}{ccccc}
        \hline\hline
        & \multicolumn{2}{c}{\#Inputs and outputs for all layers} & \multicolumn{2}{c}{\#Parameters}
        \\
        Model & w/o masking & w/ masking & w/o masking & w/ masking
        \\
        \hline
        Dynamics & 302 & 92 & 5,810 & 1,526
        \\
        Reward & 155 & 71 & 3,752 & 1,232
        \\
        \hline
        Total & 457 & 163 & 9,562 & 2,758
        \\
        \hline\hline
    \end{tabular}
\end{table}

\begin{figure*}[tb]
    \begin{subfigure}[b]{0.48\linewidth}
        \centering
        \includegraphics[keepaspectratio=true,width=\linewidth]{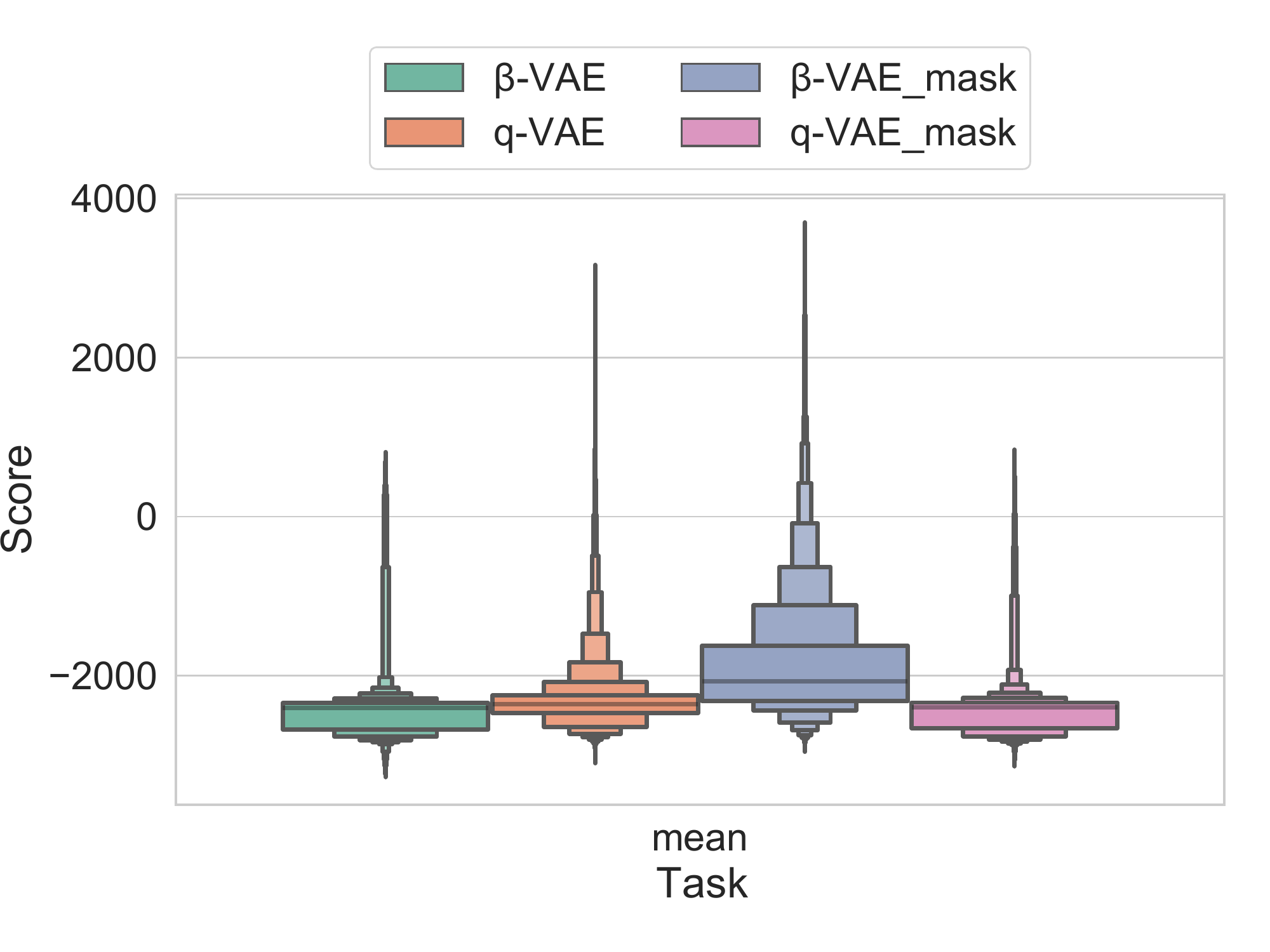}
        \subcaption{Dynamics}
        \label{fig:sim_world_dyn}
    \end{subfigure}
    \begin{subfigure}[b]{0.48\linewidth}
        \centering
        \includegraphics[keepaspectratio=true,width=\linewidth]{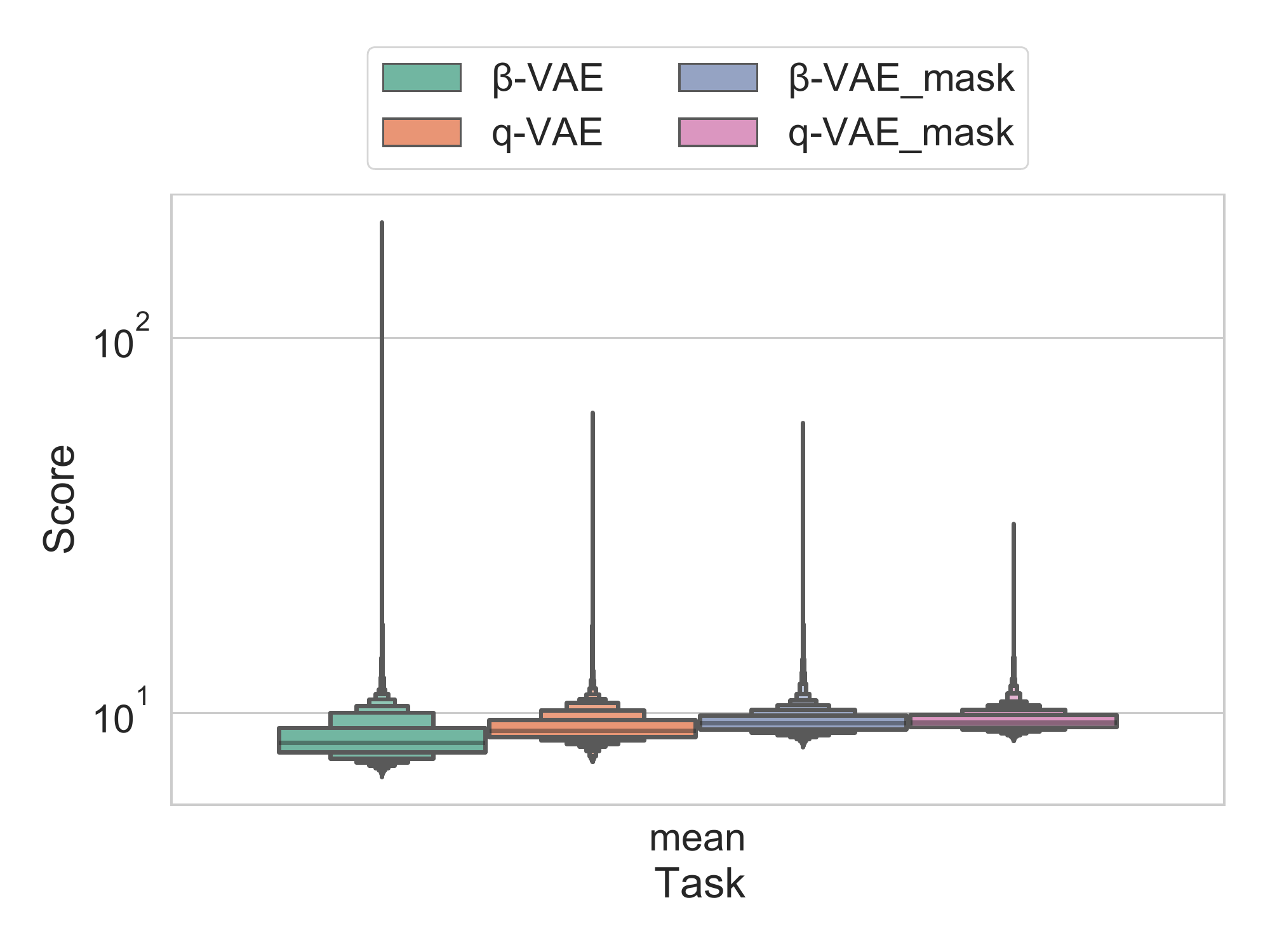}
        \subcaption{Reward}
        \label{fig:sim_world_rew}
    \end{subfigure}
    \caption{Negative log-likelihood of the world model:
        the masked q-VAE achieved the stable prediction performance, while others encountered the worst-case prediction error;
        in particular, the masked $\beta$-VAE failed to learn the dynamics since it lost the essential information for predicting future states.
    }
    \label{fig:sim_world}
\end{figure*}

The control performance by CEM is compared between $\beta$-VAE and q-VAE, both of which obtained the sufficiently meaningful latent space.
Here, $|\mathcal{Z}|$ was set to be 50, which cannot reduce the computational cost sufficiently.
Therefore, in order to confirm the benefit of sparsification, the unnecessary latent dimensions are excluded by masking, and the state space with minimal realization is extracted.
As a criterion for judging unnecessary dimensions, the sample standard deviation of the locations of the encoder is evaluated (see Fig.~\ref{fig:sim_latent}).
The dimension with the small sample standard deviation mostly takes zero for most of the data, and can be eliminated as an unnecessary dimension.
As can be seen in the figure, it is easy to expect that the top eight dimensions (with a standard deviation over 0.15) are important in the case of q-VAE.
In line with this, $\beta$-VAE also extracts the top eight dimensions, but there is concern that the necessary information may be truncated.

The world model constructed on the latent space before and after masking is used to implement control with CEM.
Each method was tested with different random seeds for 300 episodes, and if no failure occurred during the episode, it was successfully completed after 200 steps.
The statistical number of steps and the number of CEM iterations in each episode are depicted in Fig.~\ref{fig:sim_ctrl}.
It is remarkable that masking reduced the computational cost and increased the number of iterations.
In fact, the number of inputs/outputs and parameters of the world model are reduced by masking, as shown in Table~\ref{tab:size}.

The number of steps reveals the performance difference between the methods.
First, $\beta$-VAE clearly reduced the maximum performance by masking.
This is because the necessary information was missing due to masking, and the optimization by CEM did not work properly.
On the other hand, in q-VAE, masking improved the maximum performance more than in the other cases.
One of the reason for this is simply that the necessary information is retained even after masking, facilitating the optimization by increasing the number of iterations.

In addition, it was confirmed that q-VAE without masking requires a larger learning rate than the others (from $10^{-3}$ to $10^{-2}$) for making learning of the world model progress.
This may be due to the fact that most of inputs and outputs in the training dataset were zero, resulting in over-learning that generate zero.
Therefore, the performance of q-VAE without masking may have been insufficient.
In fact, the negative log-likelihood of the world model for the test dataset, as shown in Fig.~\ref{fig:sim_world}, revealed that q-VAE without masking increased the worst-case loss of the dynamics model.
Although the learning rate could be adjusted to make progress, over-learning still occurred in favor of majority zeros, overlooking the important features.
Note that the masked $\beta$-VAE was reduced the accuracy of the dynamics model, as expected above.

\section{Experiment}

\subsection{Task}

\begin{figure*}[tb]
    \begin{subfigure}[b]{0.48\linewidth}
        \centering
        \includegraphics[keepaspectratio=true,width=\linewidth]{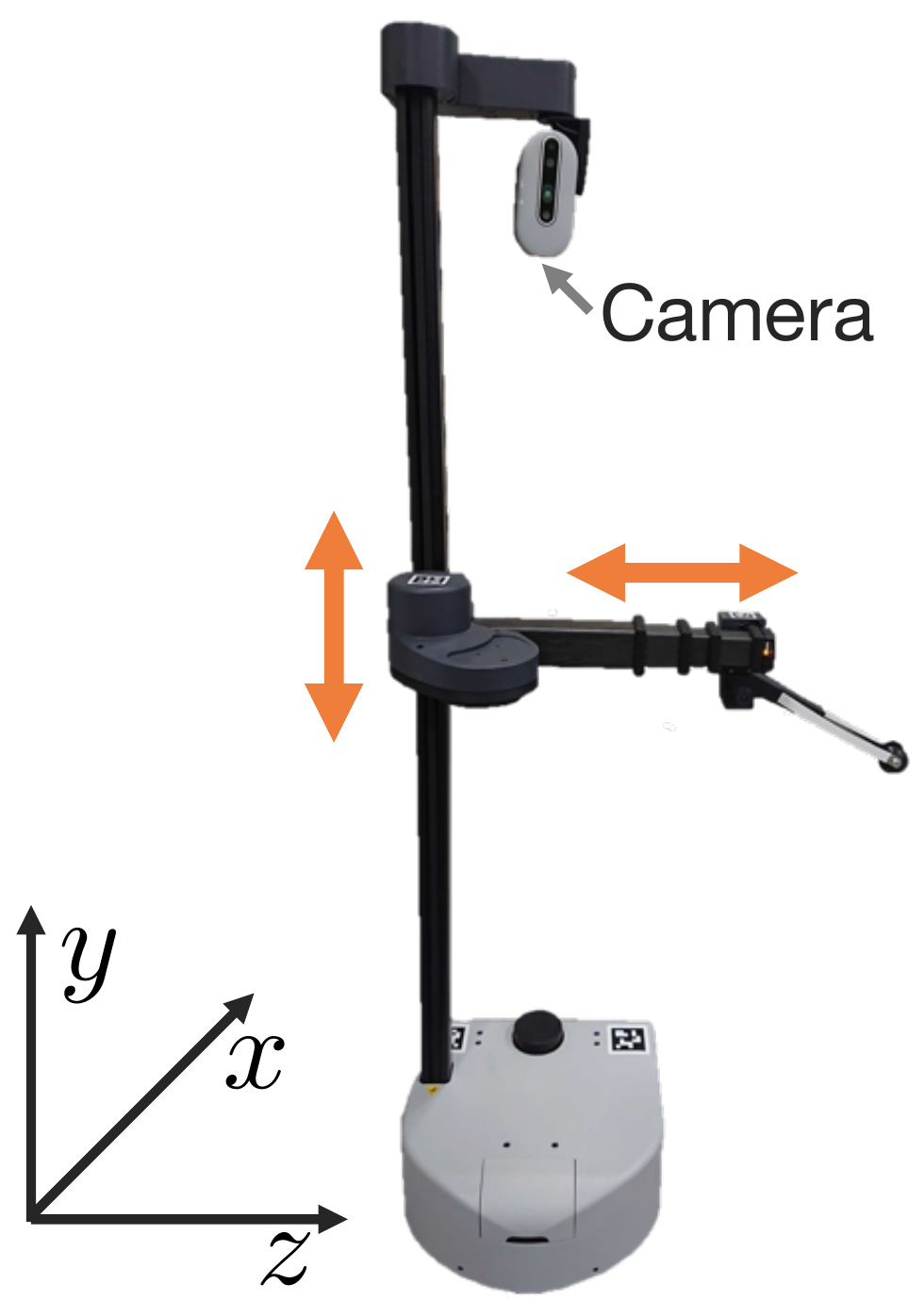}
        \subcaption{Stretch RE1}
        \label{fig:exp_env_stretch}
    \end{subfigure}
    \begin{subfigure}[b]{0.48\linewidth}
        \centering
        \includegraphics[keepaspectratio=true,width=\linewidth]{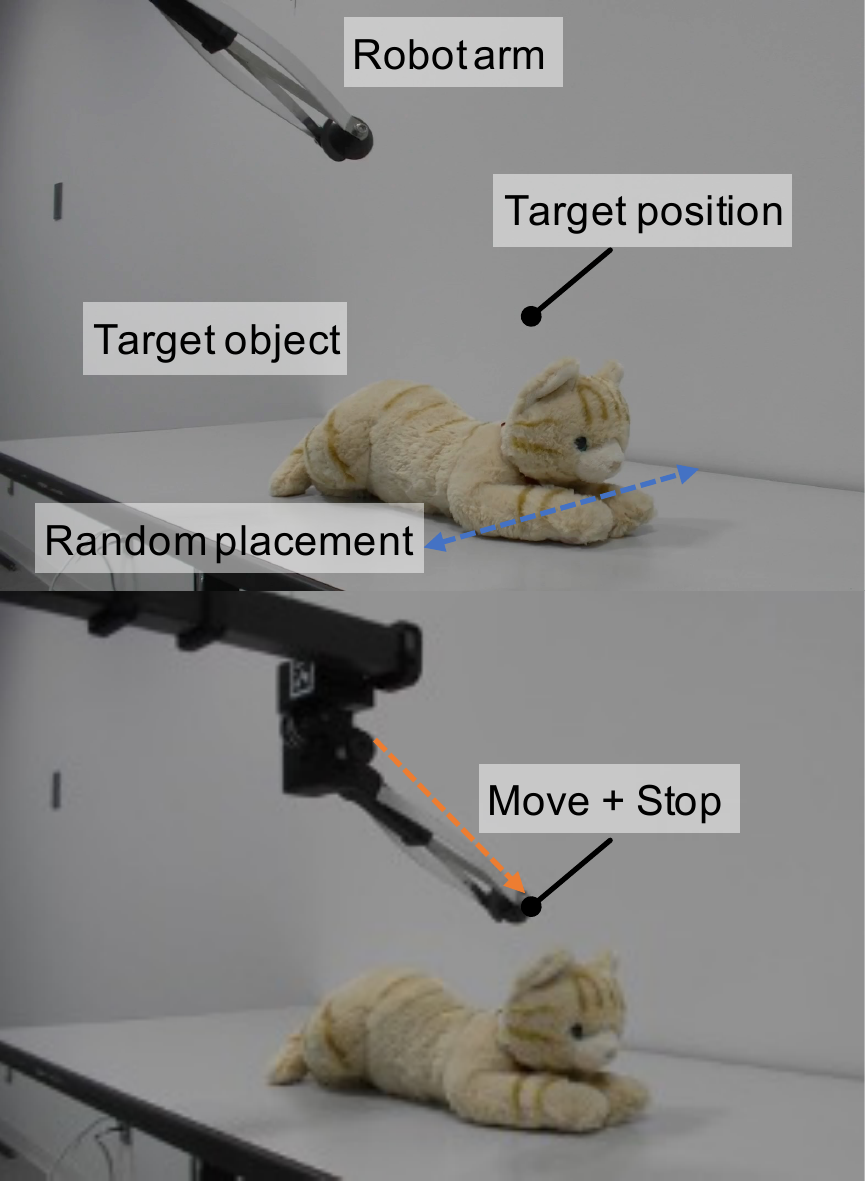}
        \subcaption{Task}
        \label{fig:exp_env_task}
    \end{subfigure}
    \caption{Experimental setup:
        (a) this task limits the motion of Stretch RE1 to $yz$-axes arm movement by two linear actuations, while the task scene is observed by a camera on its top;
        (b) this task lets the tip of arm reach a target position, 2~cm above a target object (and around its head).
    }
    \label{fig:exp_env}
\end{figure*}

As a demonstration, we conduct a reaching task, named \textit{stretch-reach}, with a Stretch RE1 developed by Hello Robot~\cite{kemp2022design}.
Specifically, as shown in Fig.~\ref{fig:exp_env_stretch}, Stretch RE1 is a kind of mobile manipulator with a camera on its top.
For simplicity, the motion of this robot is limited to $yz$-axes arm movement (within $[0.87, 1]$ in $y$ direction and $[0, 0.5]$ in $z$ direction).
The target position is 2~cm above an object randomly placed in the $z$ direction (within $[0.15, 0.45]$), and the task is to move the arm to that position (see Fig.~\ref{fig:exp_env_task}).
Similar to the above simulation, one of the observations is an RGB image (originally, 424$\times$240$\times$3), which is pre-processed to 64$\times$64$\times$3.
In addition, since the observation of the arm position and velocity can be easily measured from the encoders for the respective actuators, this 4D information is also added as an observation.

The task is considered successful when the arm stops near the target position.
Specifically, the following reward function $r$ is provided, and $r \geq -0.02$ is considered success, and 20 steps without success is considered failure.
\begin{align}
    r = - (e + 0.3v)
\end{align}
where $e$ denotes the distance between the current arm position and the target position, and $v$ denotes the arm velocity.
From the above setup, although the task itself is simple, the random target position has to be estimated from the RGB image to compute $e$ and to obtain the world model.

Note that the details of other configurations are described in Appendix~\ref{app:config}, similar to that for the simulation.

\subsection{Results}

\begin{table}[tb]
    \caption{Comparisons for \textit{stretch-reach}}
    \label{tab:comp_stretch}
    \centering
    \begin{tabular}{cccccccc}
        \hline\hline
        Method & $q$ & $q_1$ & $q_2$ & $\zeta_1$ & $\zeta_2$ & $\beta$ & $\gamma$
        \\
        \hline
        $\beta$-VAE & 1 & -- & -- & 50 & 1 & 0.3 & --
        \\
        q-VAE (proposal) & 0.95 & 0.95 & 0.999 & 50 & 1 & 50 & 3
        \\
        \hline\hline
    \end{tabular}
\end{table}

\begin{figure*}[tb]
    \begin{subfigure}[b]{0.48\linewidth}
        \centering
        \includegraphics[keepaspectratio=true,width=\linewidth]{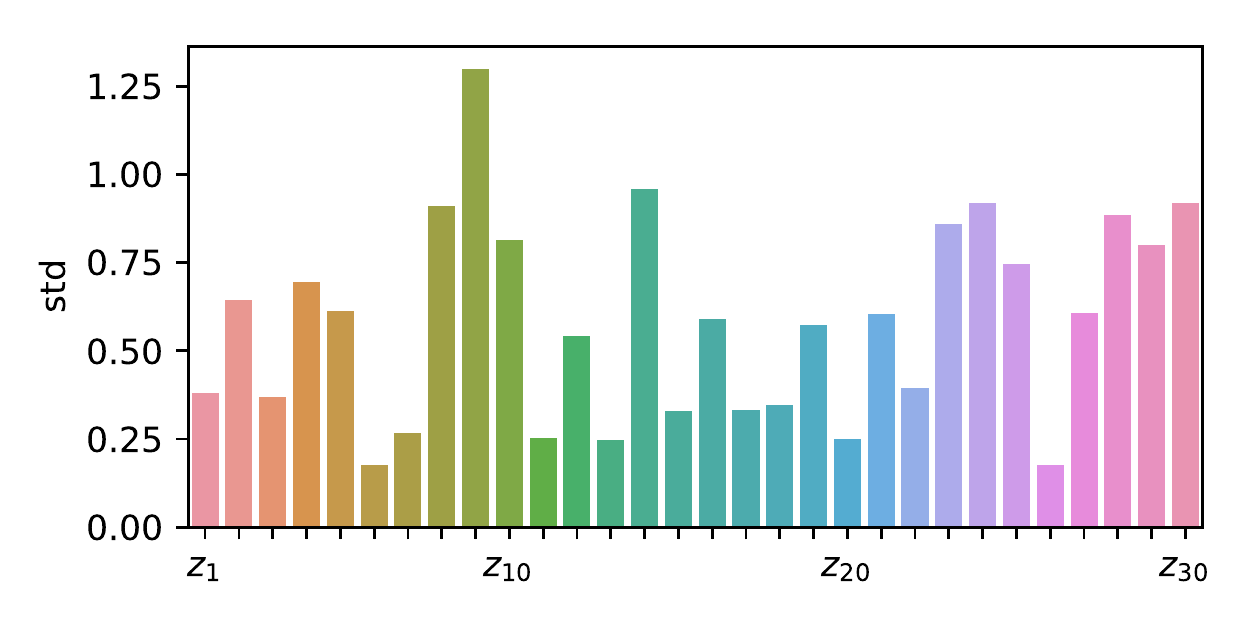}
        \subcaption{$\beta$-VAE}
        \label{fig:exp_latent_bvae}
    \end{subfigure}
    \begin{subfigure}[b]{0.48\linewidth}
        \centering
        \includegraphics[keepaspectratio=true,width=\linewidth]{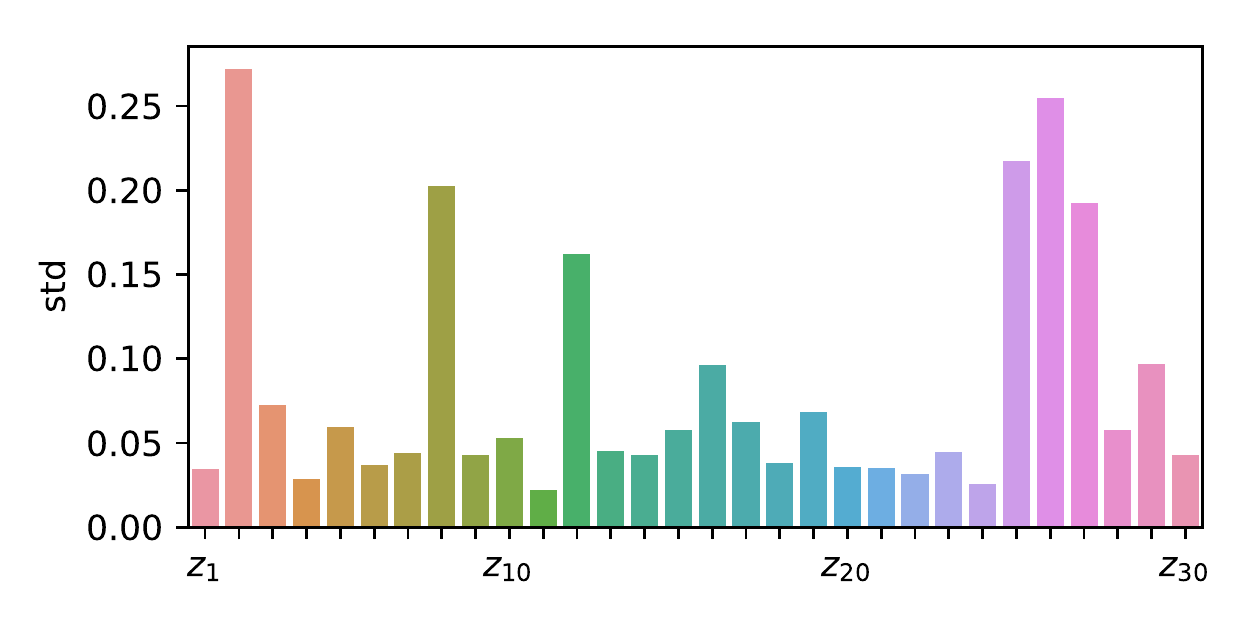}
        \subcaption{q-VAE}
        \label{fig:exp_latent_qvae}
    \end{subfigure}
    \caption{Importance of the latent dimensions:
        as well as the simulation shown in Fig.~\ref{fig:sim_latent}, only q-VAE succeeded in easily selecting the important latent dimensions.
    }
    \label{fig:exp_latent}
\end{figure*}

Since the standard VAE is insufficient to extract the important information, only two methods, $\beta$-VAE and q-VAE, are tested as described in Table~\ref{tab:comp_stretch}.
Note again that q-VAE is set to satisfy the sparsity condition indicated by eq.~\eqref{eq:condition} while confirming the stability of the numerical computation.

As in the simulation, after confirming that the reconstruction accuracy was comparable to each other, the importance of the latent dimensions was evaluated in terms of the sample standard deviation (see Fig.~\ref{fig:exp_latent}).
From the figure, q-VAE can select six dimensions (with the same threshold 0.15).
In theory, this task may have five dimensions for the minimal realization (i.e. the 2D arm position and velocity and the $z$-axis target position), but in reality, other environmental noise (e.g. misalignment of the target object in the $x$ direction) may occur.
Therefore, a total of six dimensions can be reasonably extracted: five dimensions for the theoretical minimal realization and one dimension to handle other noise in practice.
On the other hand, $\beta$-VAE is difficult to find the important dimensions.

For practical use, the size of dimensions extracted by q-VAE is unknown when using $\beta$-VAE alone, so masking to $\beta$-VAE is omitted in this experiment.
In addition, the simulation results described before indicated that q-VAE without masking is prone to make learning of the world model unstable by unnecessary axes, so we omit it as well.
Therefore, the performance comparison in this experiment is limited to $\beta$-VAE without masking and q-VAE with masking.

\subsection{Accuracy of world model}

\begin{figure*}[tb]
    \begin{subfigure}[b]{0.48\linewidth}
        \centering
        \includegraphics[keepaspectratio=true,width=\linewidth]{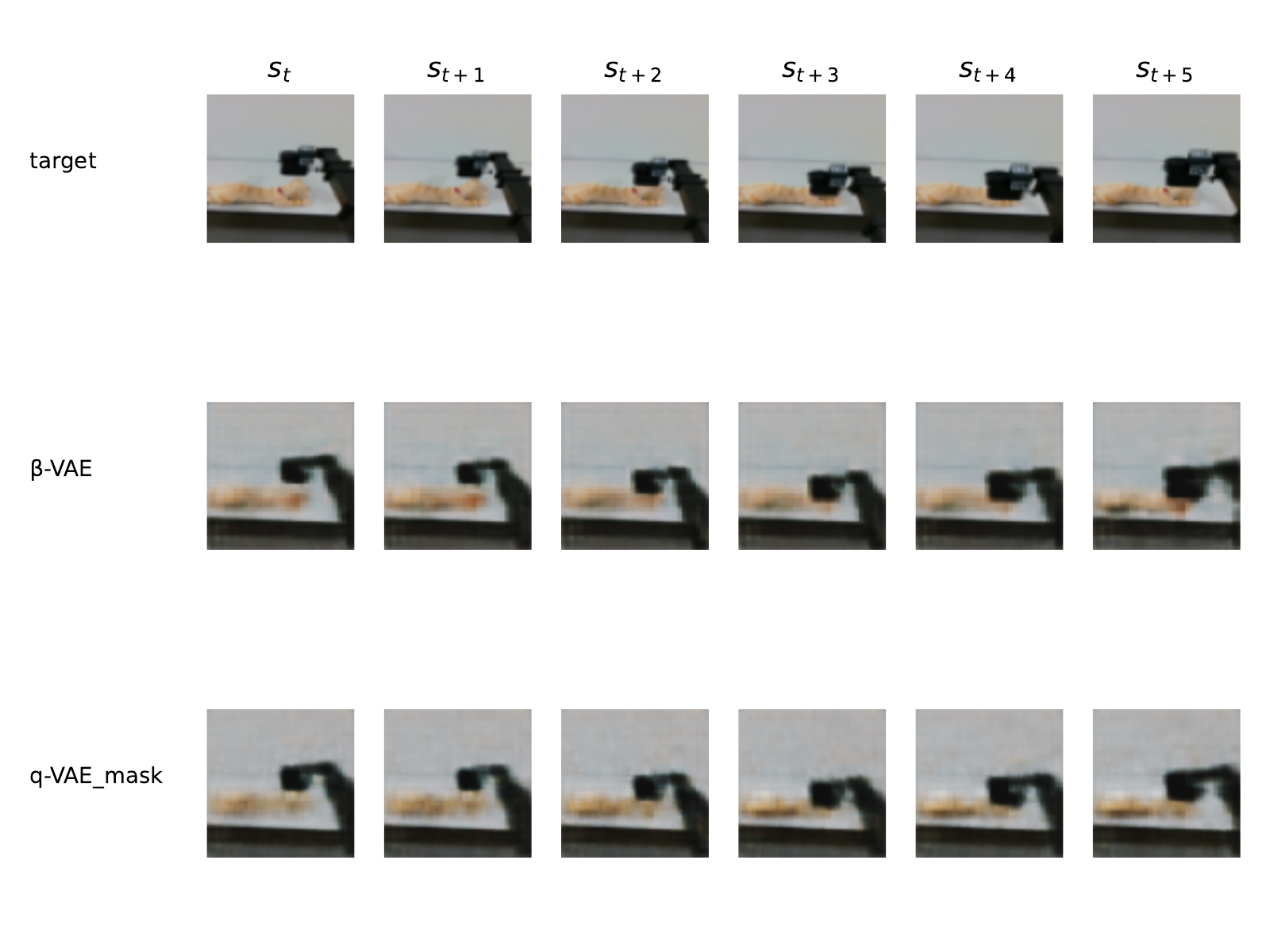}
        \subcaption{Image $x_{\mathrm{img}}$}
        \label{fig:exp_pred_img}
    \end{subfigure}
    \begin{subfigure}[b]{0.48\linewidth}
        \centering
        \includegraphics[keepaspectratio=true,width=\linewidth]{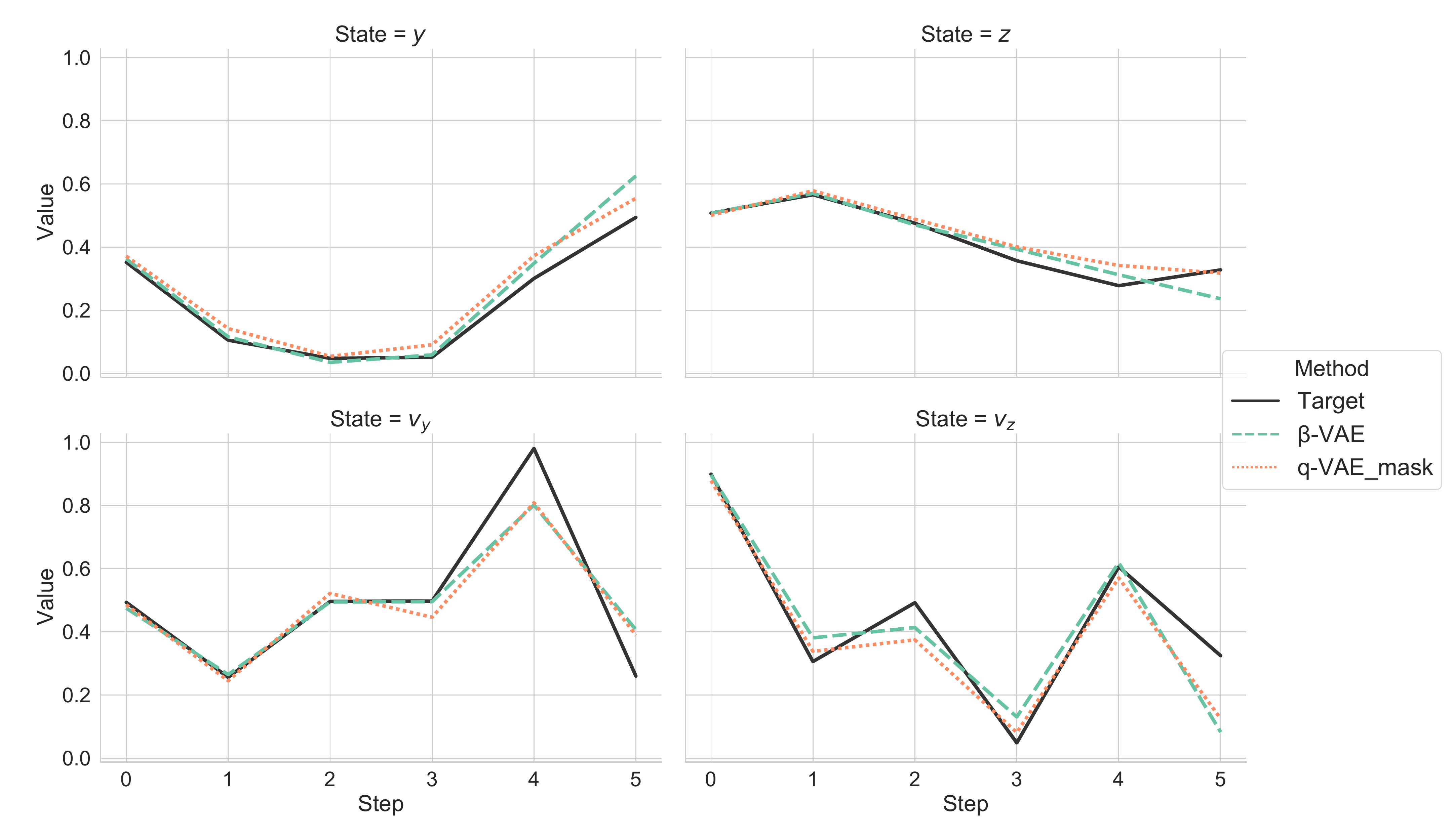}
        \subcaption{Arm state $x_{\mathrm{arm}}$}
        \label{fig:exp_pred_arm}
    \end{subfigure}
    \caption{Prediction of future observations ($H=5$ steps ahead):
        in both (a) $x_{\mathrm{img}}$ and (b) $x_{\mathrm{arm}}$, the masked q-VAE was comparable to $\beta$-VAE.
    }
    \label{fig:exp_pred}
\end{figure*}

\begin{figure}[tb]
    \centering
    \includegraphics[keepaspectratio=true,width=0.96\linewidth]{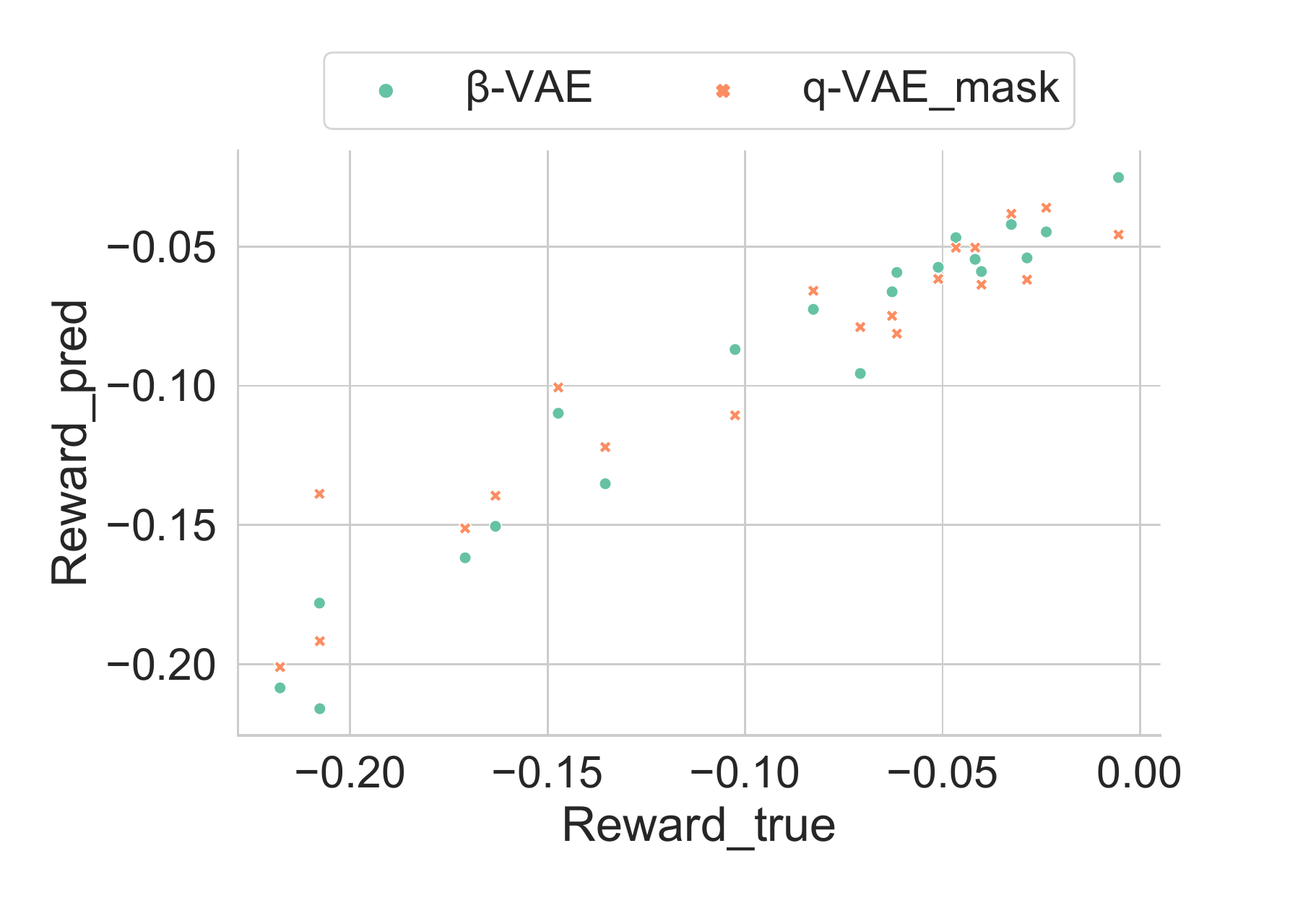}
    \caption{Accuracy of the reward prediction:
        both methods could approximate the reward function.
    }
    \label{fig:exp_reward}
\end{figure}

First, we confirm that the performance of the world models is comparable to each other.
The dynamics to $H=5$ horizon ahead handled in CEM is illustrated in Fig.~\ref{fig:exp_pred}.
It can be seen that the differences between the predictions and the true observations for $\beta$-VAE and q-VAE are comparable for both the image and the arm state.
The predictions and true values of the rewards are also compared in Fig.~\ref{fig:exp_reward}.
Similarly, the predictions are mostly consistent with the true values, suggesting that both methods achieved the good accuracy.
From the above, it can be expected that the difference in control performance between the two methods (to be confirmed in the next section) would be only due to the masking (from 30 to six dimensions) obtained by the sparsity of q-VAE.

\subsection{Control performance}

\begin{figure*}[tb]
    \begin{subfigure}[b]{0.32\linewidth}
        \centering
        \includegraphics[keepaspectratio=true,width=\linewidth]{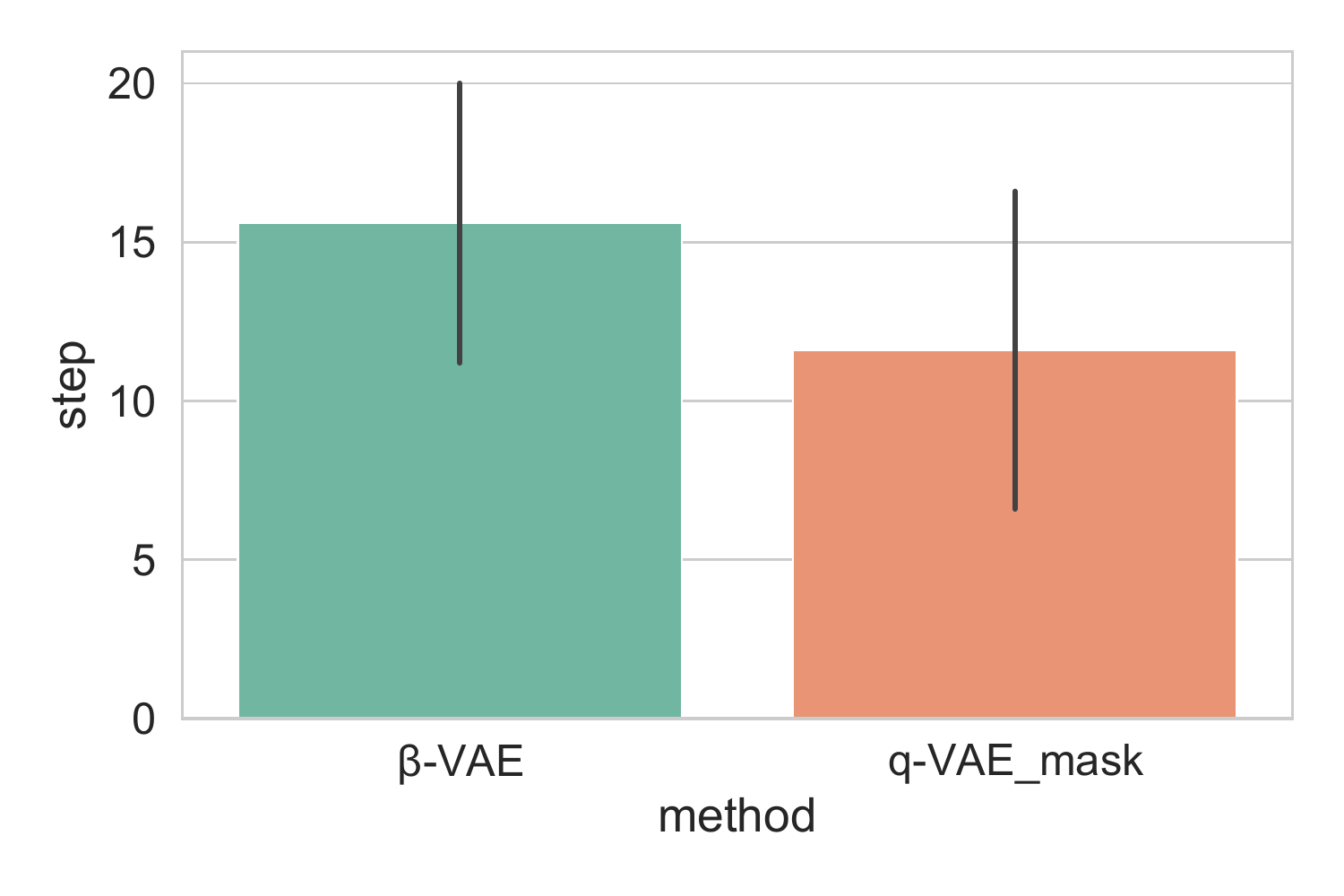}
        \subcaption{\#Step for 0.2~m}
        \label{fig:exp_result_02_step}
    \end{subfigure}
    \begin{subfigure}[b]{0.32\linewidth}
        \centering
        \includegraphics[keepaspectratio=true,width=\linewidth]{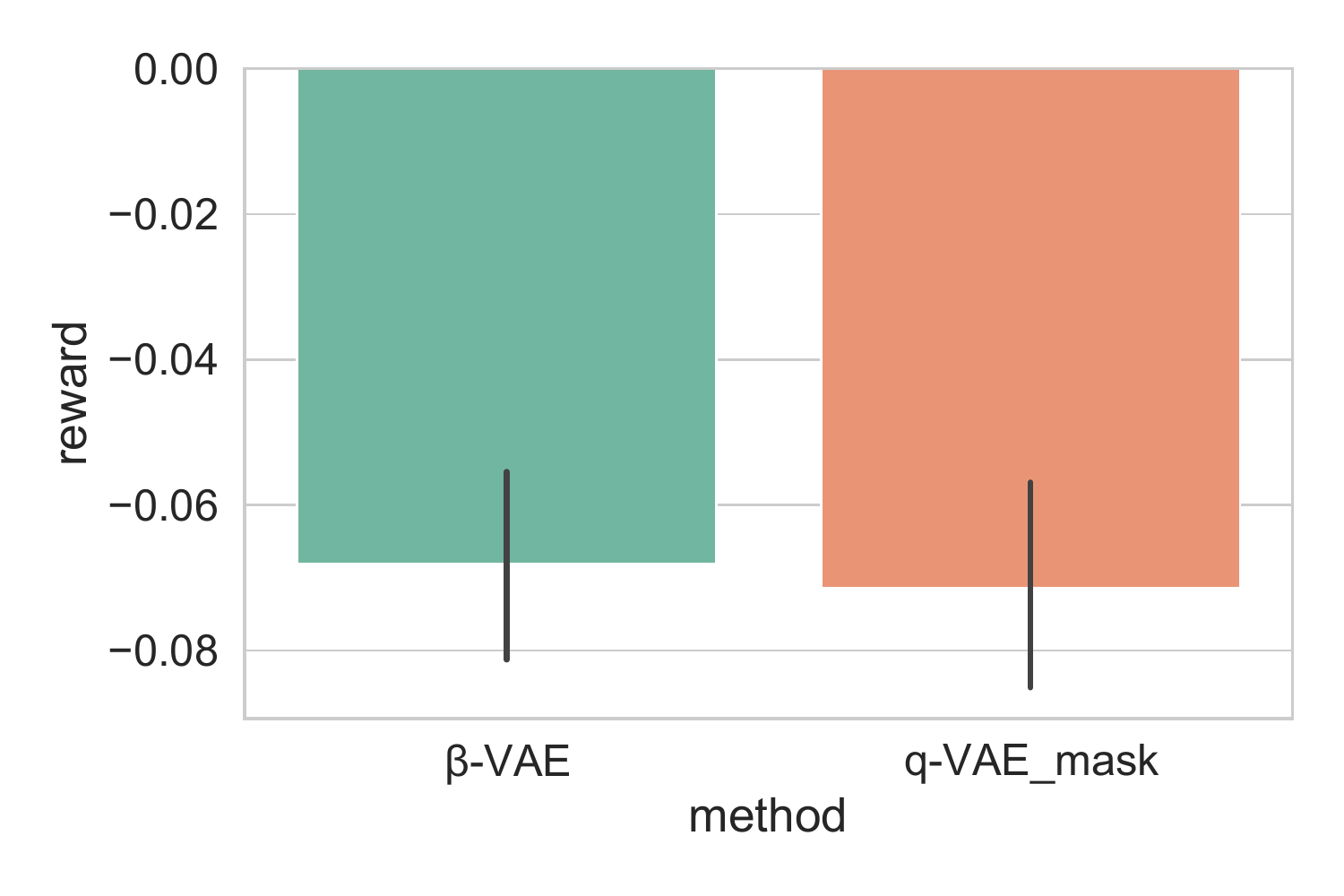}
        \subcaption{Reward for 0.2~m}
        \label{fig:exp_result_02_rew}
    \end{subfigure}
    \begin{subfigure}[b]{0.32\linewidth}
        \centering
        \includegraphics[keepaspectratio=true,width=\linewidth]{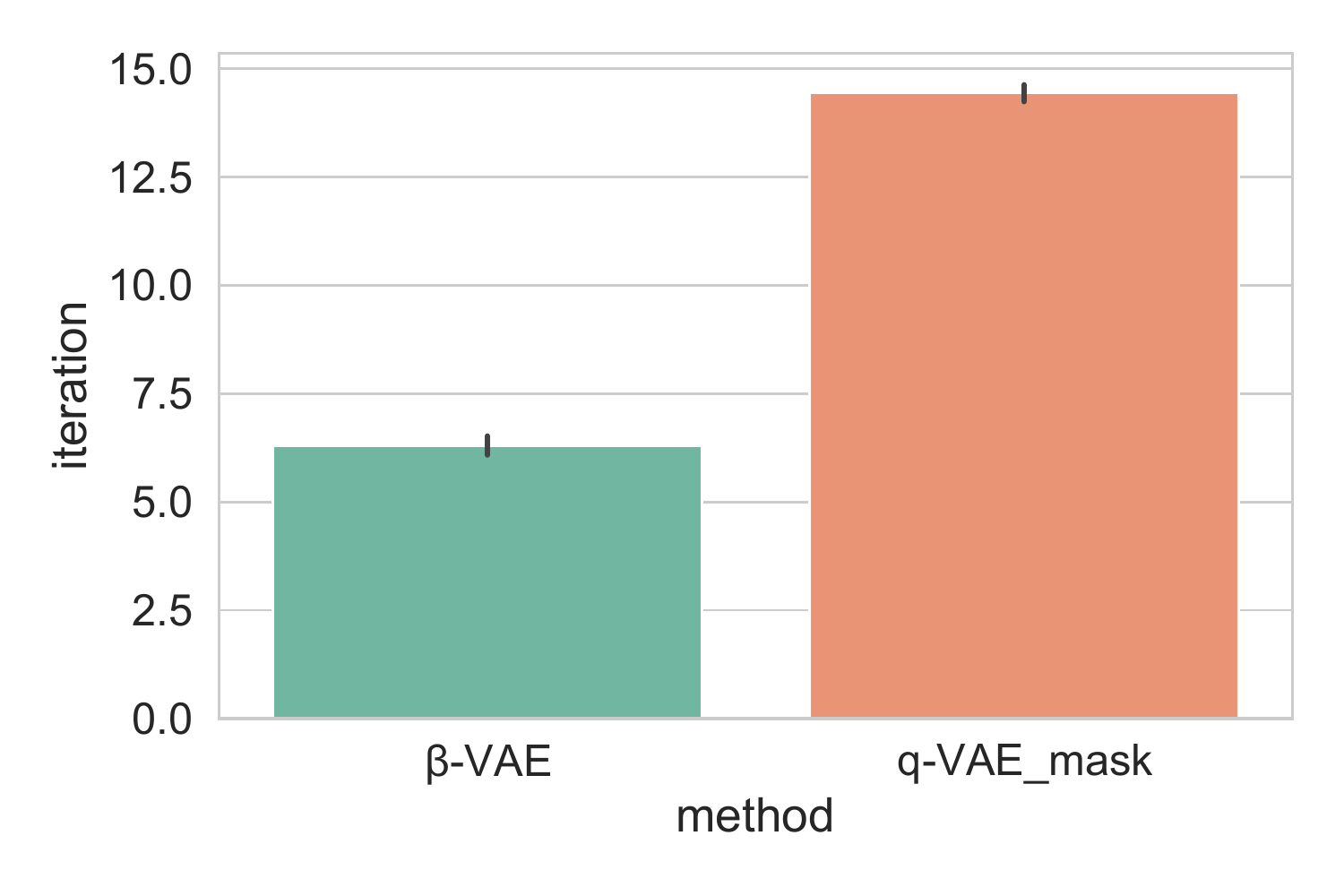}
        \subcaption{\#Iteration for 0.2~m}
        \label{fig:exp_result_02_iter}
    \end{subfigure}
    \begin{subfigure}[b]{0.32\linewidth}
        \centering
        \includegraphics[keepaspectratio=true,width=\linewidth]{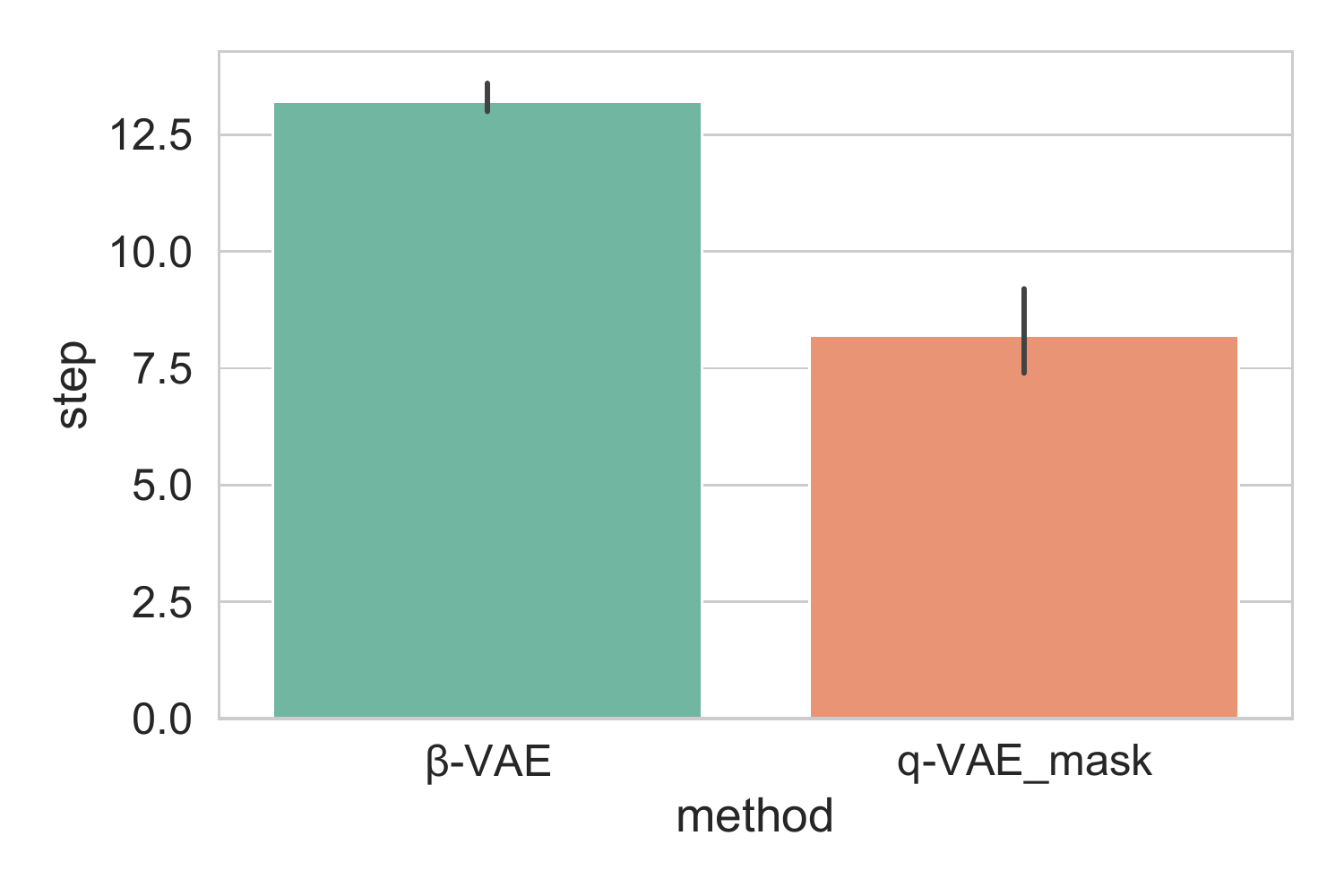}
        \subcaption{\#Step for 0.3~m}
        \label{fig:exp_result_03_step}
    \end{subfigure}
    \begin{subfigure}[b]{0.32\linewidth}
        \centering
        \includegraphics[keepaspectratio=true,width=\linewidth]{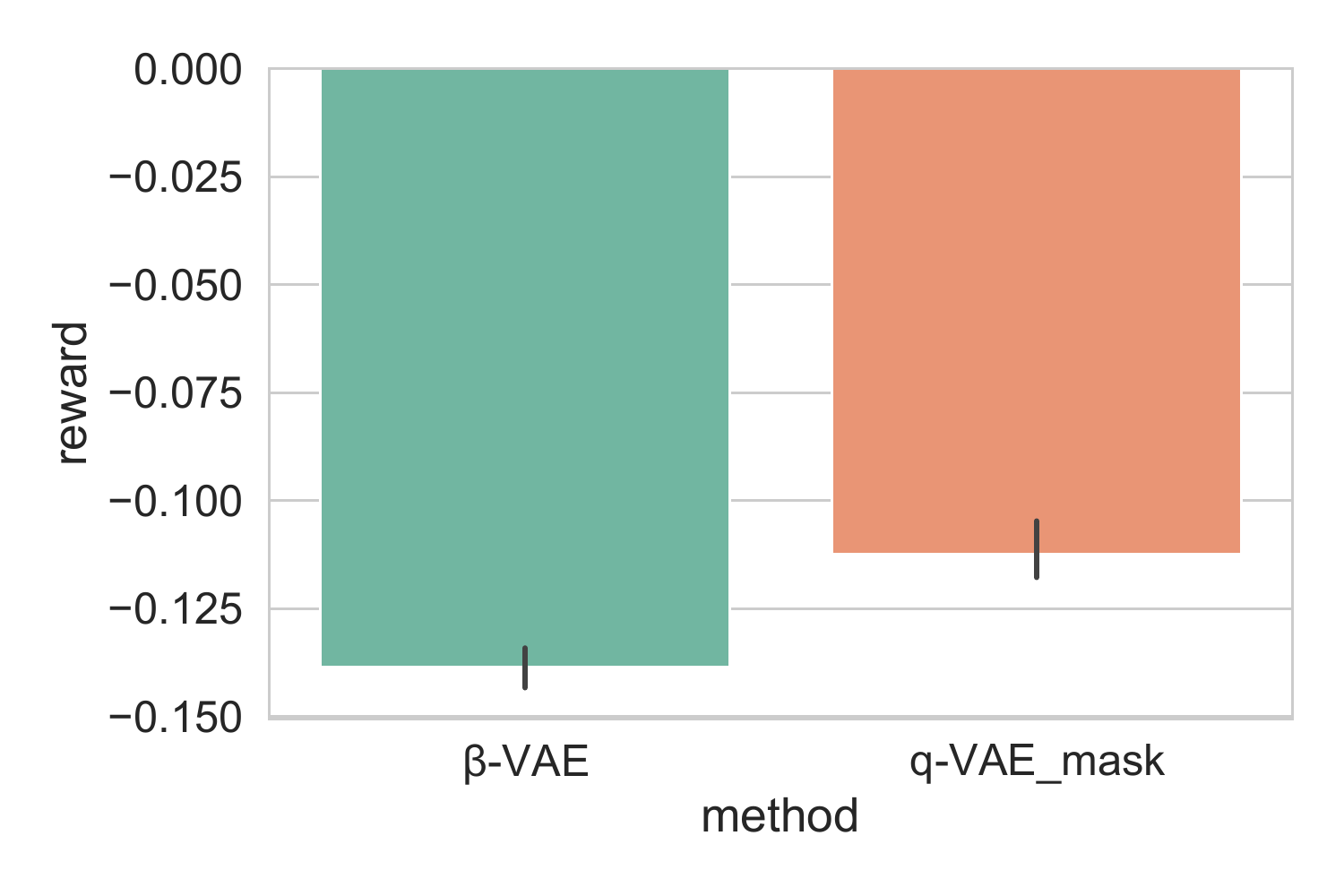}
        \subcaption{Reward for 0.3~m}
        \label{fig:exp_result_03_rew}
    \end{subfigure}
    \begin{subfigure}[b]{0.32\linewidth}
        \centering
        \includegraphics[keepaspectratio=true,width=\linewidth]{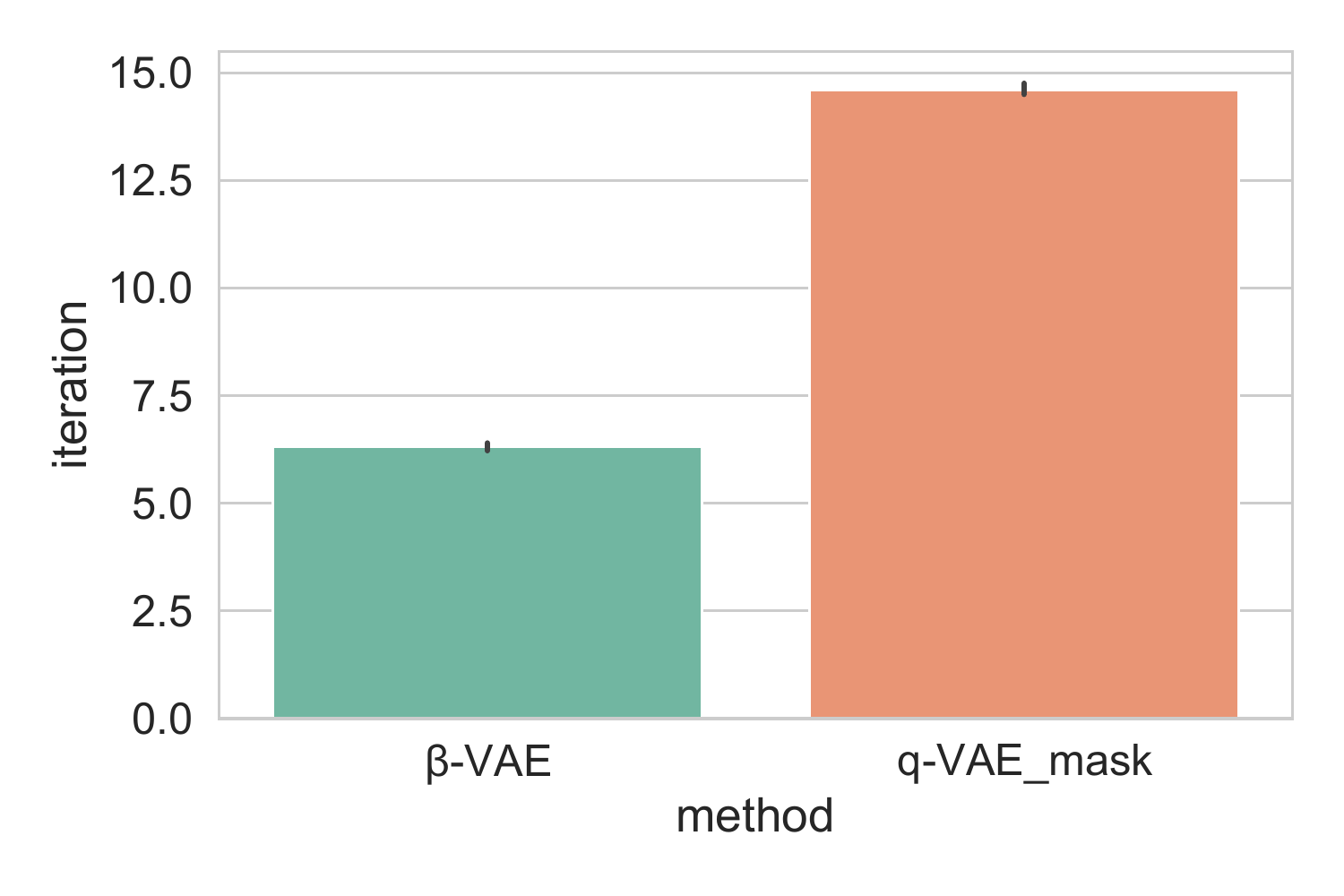}
        \subcaption{\#Iteration for 0.3~m}
        \label{fig:exp_result_03_iter}
    \end{subfigure}
    \begin{subfigure}[b]{0.32\linewidth}
        \centering
        \includegraphics[keepaspectratio=true,width=\linewidth]{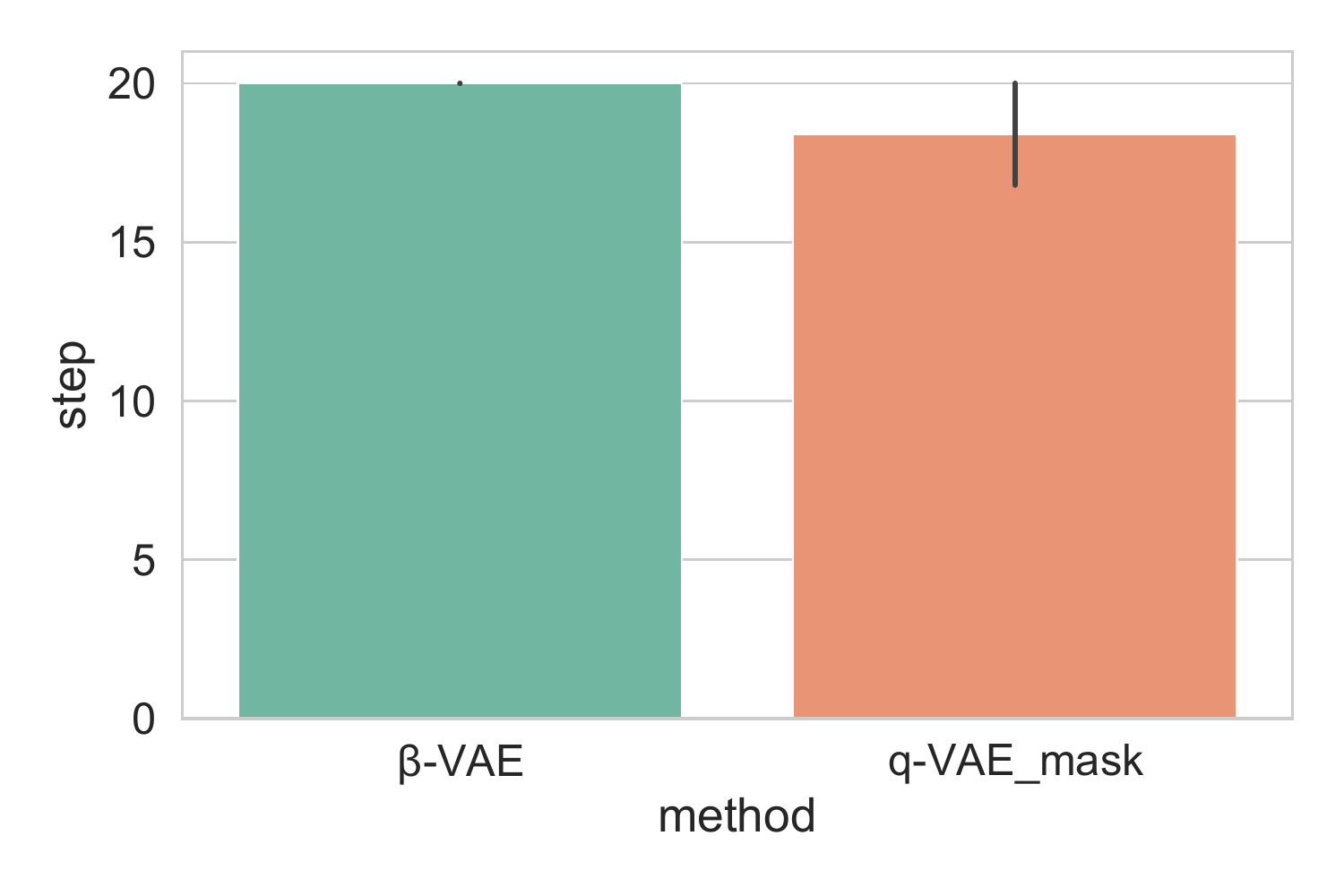}
        \subcaption{\#Step for 0.4~m}
        \label{fig:exp_result_04_step}
    \end{subfigure}
    \begin{subfigure}[b]{0.32\linewidth}
        \centering
        \includegraphics[keepaspectratio=true,width=\linewidth]{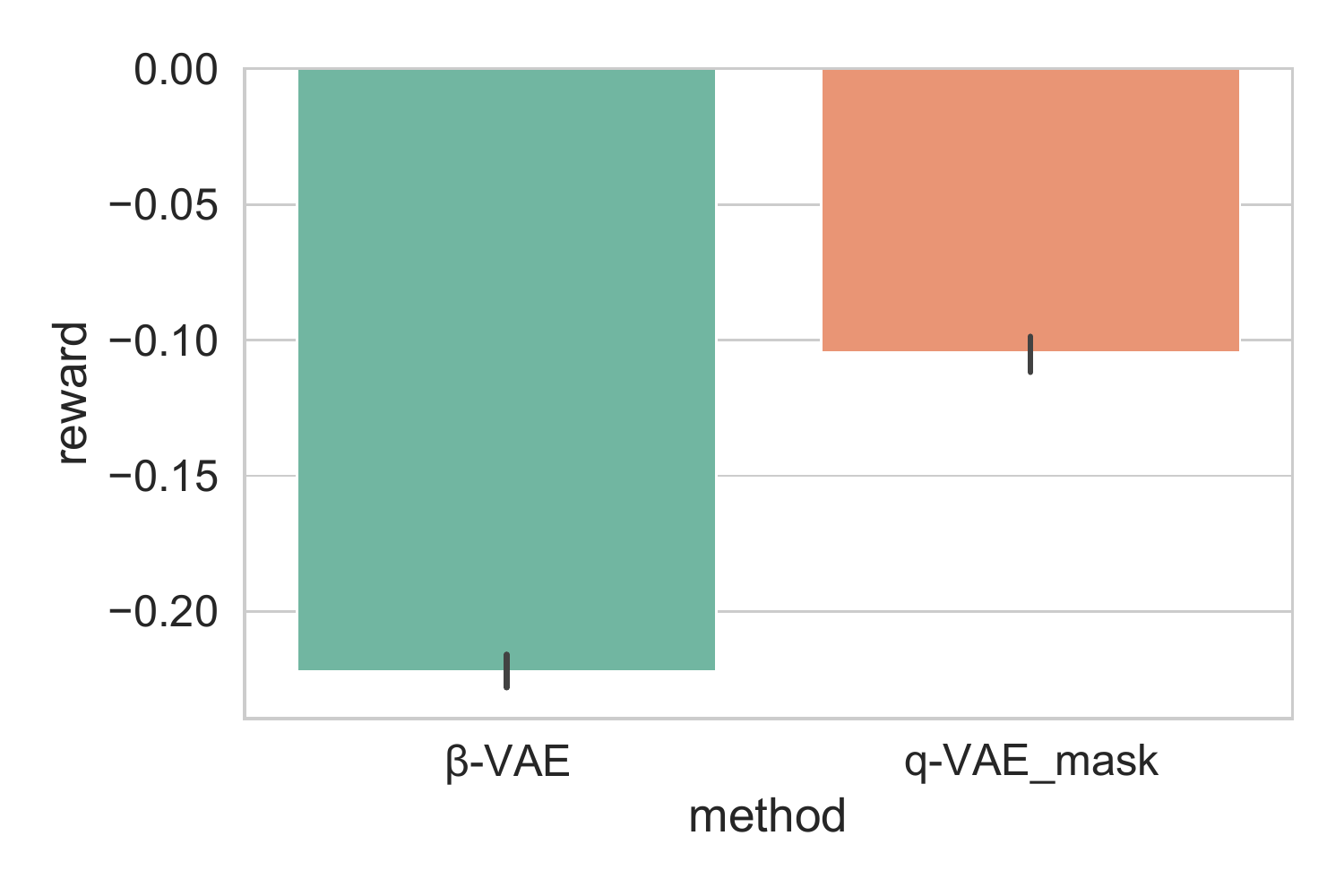}
        \subcaption{Reward for 0.4~m}
        \label{fig:exp_result_04_rew}
    \end{subfigure}
    \begin{subfigure}[b]{0.32\linewidth}
        \centering
        \includegraphics[keepaspectratio=true,width=\linewidth]{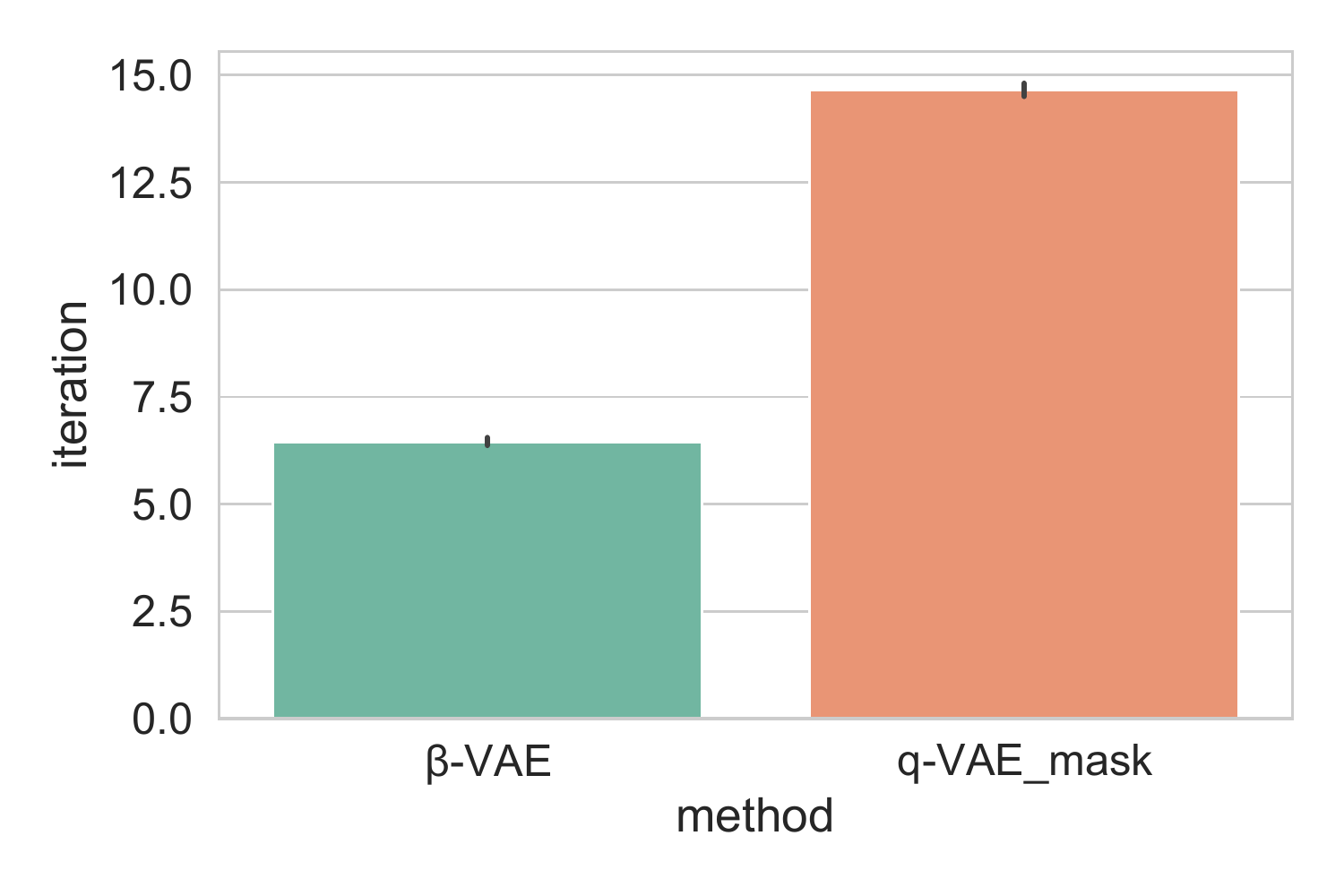}
        \subcaption{\#Iteration for 0.4~m}
        \label{fig:exp_result_04_iter}
    \end{subfigure}
    \caption{Experimental results:
        in (c), (f), (i), as well as the simulation result shown in Fig.~\ref{fig:sim_ctrl}, the number of iterations was able to be increased by masking;
        the reduction in computational cost and the increase in the number of iterations due to masking contributed significantly to control performance, accelerating the time to reach the target position and/or increasing the sum of rewards.
    }
    \label{fig:exp_result}
\end{figure*}

\begin{figure}[tb]
    \centering
    \includegraphics[keepaspectratio=true,width=0.96\linewidth]{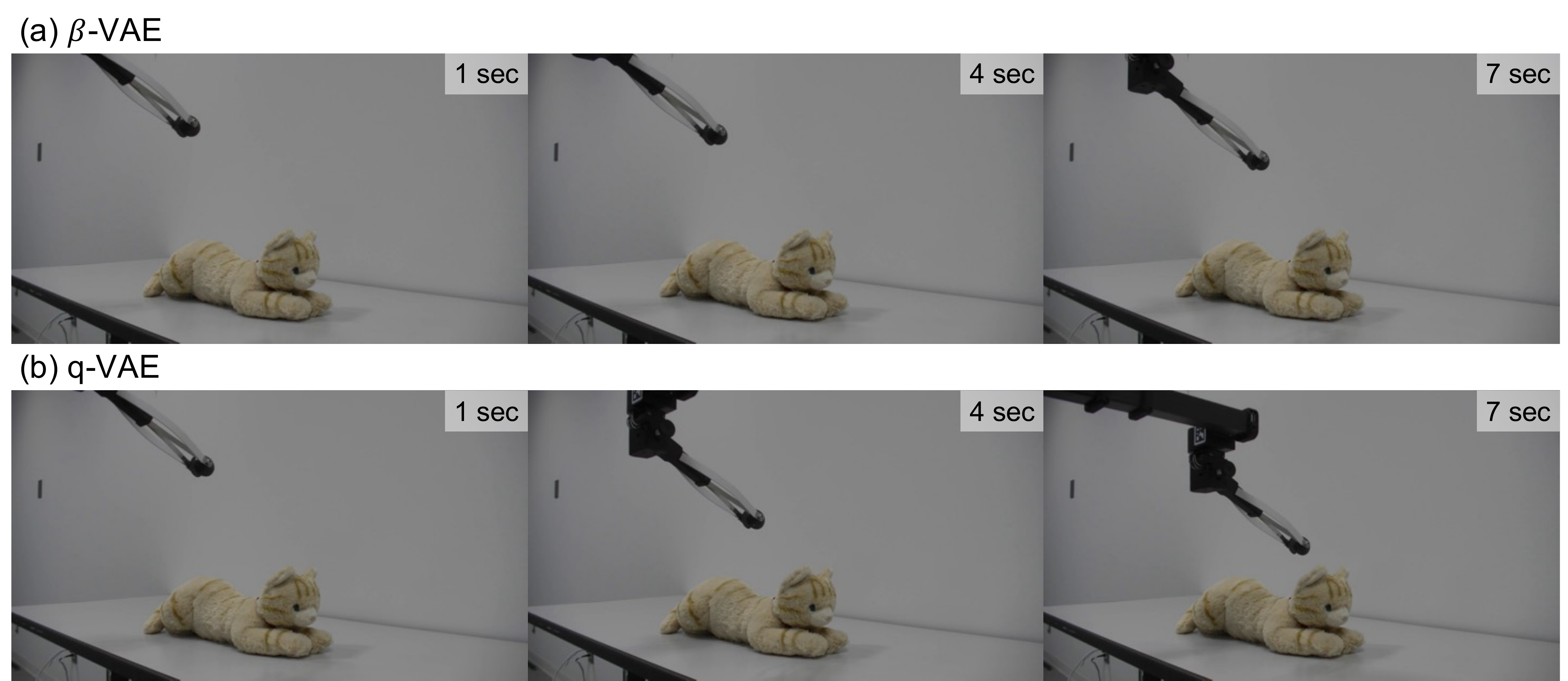}
    \caption{Snapshots of acquired motions:
        the masked q-VAE made the tip of arm reach the target position faster than $\beta$-VAE.
    }
    \label{fig:exp_snap}
\end{figure}

Using the learned world model, 5 trials of reaching were performed for each of the 3 target object positions (i.e. 0.2, 0.3, and 0.4~m, respectively).
The number of steps until task termination (20 steps at maximum), average reward, and average number of iterations are listed up in Fig.~\ref{fig:exp_result}.
An example of the trials are shown in Fig.~\ref{fig:exp_snap} and the attached video.

When the object was placed at 0.2~m, q-VAE succeeded in reaching the target earlier than $\beta$-VAE, although there is little difference in reward due to the overall smaller distance penalty.
Even in the case with 0.3~m target object position, q-VAE succeeded in performing the task earlier than $\beta$-VAE, and also increased the reward due to smoother acceleration/deceleration.
When the target object was placed as far as 0.4~m, $\beta$-VAE failed in all trials, while q-VAE generally succeeded in most cases.

As noted above, these performance gains are not due to differences in the prediction accuracy of the world model, but rather due to the reduced computational cost by the compact world model, which is with almost the minimal realization.
In fact, the number of iterations of q-VAE is more than double that of $\beta$-VAE under all conditions.
Thus, we confirmed that q-VAE facilitates the minimal realization of the world model through sparsification and contributes to improving the control performance of computationally expensive optimal control such as CEM in real time.

\section{Conclusion and discussion}

\subsection{Conclusion}

In this paper, we improved and analyzed q-VAE, a deep learning technique for extracting sparse latent space, aiming at the minimal realization of the world model.
In particular, we clarified the hyperparameters condition under which the modified q-VAE always sparsifies the latent space.
In both simulations and experiments, the modified q-VAE successfully collapsed many latent dimensions to zero while maintaining the same level of reconstruction accuracy as the baseline method, $\beta$-VAE, by learning according to the condition.
The world model with almost the minimal realization was obtained by masking the unnecessary latent dimensions and utilized in CEM, which is a sampling-based optimal control method.
Consequently, the optimization of CEM was facilitated by the reduction in computational cost, resulting in better control performance.

\subsection{Discussion for future work}

Two major open issues can be raised from our investigation.
One is how to adjust hyperparameters.
The hyperparameters of the modified q-VAE have increased due to the decomposition from the conventional q-VAE, and it is clear that their optimal values are task-specific, although the sparsification condition limits the degree of freedom in design of them.
In particular, $q \to 0$ is desirable to strongly promote sparsification, but as the literature~\cite{kobayashi2021towards} reported, a small $q$ would exclude many data from training as outliers.
In fact, in \textit{highway-env} simulations, we observed cases where other blue cars could not be reconstructed as in the standard VAE, depending on the value of $q_2$.
A framework that is adaptive or robust to this trade-off, such as meta-optimization~\cite{aotani2021meta} or ensemble learning with multiple combinations of hyperparameters~\cite{sagi2018ensemble}, would be useful.

Another open issue is the simultaneous learning of the latent space and the world model.
In this paper, the latent space extraction phase and the world model acquisition phase were conducted independently, giving priority to ease of analysis.
However, in order to obtain the state holding the minimal realization, it is desirable to extract the latent space by considering the world model (and controller).
Since simultaneous learning of multiple modules tends to be basically difficult, it would be important to introduce curriculum learning~\cite{graves2017automated}, etc.

Anyway, we will apply our method to more practical tasks in the near future and enhance its practicality by resolving the above-mentioned open issues.

\section*{Acknowledgement}

This work was supported by JSPS KAKENHI, Grant-in-Aid for Scientific Research (B), Grant Number JP20H04265 and JST, PRESTO Grant Number JPMJPR20C3, Japan.

\bibliographystyle{tfnlm}
\bibliography{biblio}

\appendix

\section{Learning configurations}
\label{app:config}

\begin{table}[tb]
    \caption{Network designs for VAE}
    \label{tab:net_vae}
    \centering
    \begin{tabular}{cc}
        \hline\hline
        Meaning & Setting
        \\
        \hline
        Kernel sizes for convolutional layers & [4, 3, 3, 3, 3]
        \\
        Channel sizes for convolutional layers & [8, 16, 32, 64, 128]
        \\
        Unit sizes for FC layers \textcircled{1} & [8, 32, 128]
        \\
        Unit sizes for FC layers \textcircled{2} & [100, 80, 60]
        \\
        \hline
        Kernel sizes for deconvolutional layers & [3, 3, 3, 3, 4]
        \\
        Channel sizes for deconvolutional layers & [128, 64, 32, 16, 8]
        \\
        Unit sizes for FC layers \textcircled{3} & [128, 32, 8]
        \\
        Unit sizes for FC layers \textcircled{4} & [60, 80, 100]
        \\
        \hline
        The size of latent space $|\mathcal{Z}|$ & \{50, 30\}
        \\
        Activation function for all layers & Swish~\cite{elfwing2018sigmoid}
        \\
        & + Layer normalization~\cite{ba2016layer}
        \\
        \hline\hline
    \end{tabular}
\end{table}

\begin{table}[tb]
    \caption{Network designs for world model}
    \label{tab:net_world}
    \centering
    \begin{tabular}{ccc}
        \hline\hline
        Meaning & Setting in simulation & Setting in experiment
        \\
        \hline
        Unit sizes for dynamics & [50, 50] & [10, 10]
        \\
        Unit sizes for reward & [20, 10] & [10, 10]
        \\
        \hline
        Activation function for all layers & \multicolumn{2}{c}{Tanh + Layer normalization~\cite{ba2016layer}}
        \\
        \hline\hline
    \end{tabular}
\end{table}

\begin{table}[tb]
    \caption{Learning configurations}
    \label{tab:learn}
    \centering
    \begin{tabular}{cccc}
        \hline\hline
        Meaning & Setting for VAE & Setting for dynamics & Setting for reward
        \\
        \hline
        Optimizer & t-Adam~\cite{ilboudo2020robust} & Adam~\cite{kingma2014adam} & Adam~\cite{kingma2014adam}
        \\
        Learning rate & $1 \times 10^{-4}$ & $1 \times 10^{-3}$ & $3 \times 10^{-4}$
        \\
        Batch size & 256 & 512 & 512
        \\
        \hline\hline
    \end{tabular}
\end{table}

In VAE, a continuous Bernoulli distribution~\cite{loaiza2019continuous} is employed for the image decoder $p(x_{\mathrm{img}} \mid z)$, and a diagonal Gaussian distribution is also employed for the velocity decoder $p(x_{\mathrm{vel}} \mid z)$.
From the fact that $|\mathcal{X}_{\mathrm{img}}| \gg |\mathcal{X}_{\mathrm{vel}}|$, $c=1$ in eq.~\eqref{eq:mqvae} is for $\mathcal{X}_{\mathrm{vel}}$ and $c=2$ corresponds to $\mathcal{X}_{\mathrm{img}}$.
The encoder $p(z \mid x)$ is given by a diagonal Gaussian distribution.

The details of each module in the figure are summarized in Table~\ref{tab:net_vae}.
In addition, the network architecture for the world model is also summarized in Table~\ref{tab:net_world}.
The conditions related to the learning of the respective modules are summarized in Table~\ref{tab:learn}.
In this way, after training multiple models of VAE with different random seeds, the world model is trained by selecting the median model among them.

\subsection{Configurations for simulation}

\begin{table}[tb]
    \caption{Configurations for \textit{highway-env} and CEM}
    \label{tab:param_highway}
    \centering
    \begin{tabular}{ccc}
        \hline\hline
        Symbol & Meaning & Value
        \\
        \hline
        $|\mathcal{A}|$ & The size of action space & 2
        \\
        $|\mathcal{X}_{\mathrm{img}}|$ & The size of image observation & 64$\times$64$\times$3
        \\
        $|\mathcal{X}_{\mathrm{vel}}|$ & The size of velocity observation & 2
        \\
        -- & Control frequency & 10~Hz
        \\
        $H$ & Horizon step & 10
        \\
        $K$ & The number of candidates & 10,000
        \\
        -- & Maximum iteration & 10
        \\
        $\nu$ & Elite ratio & 0.01
        \\
        $\eta$ & Smooth update & 0.4
        \\
        \hline\hline
    \end{tabular}
\end{table}

The network architecture of VAE implemented by PyTorch~\cite{paszke2017automatic} is illustrated in Fig.~\ref{fig:impl_vae}.
Since the simulation task is more complex than the experimental one, the latent dimension size $|\mathcal{Z}|$ is given a larger value of 50, and the world model is also designed slightly larger than in the experiment (although still small enough).

The control frequency is set to 10~Hz.
Because it is necessary to predict some time ahead for safe driving, Horizon set to be $H=10$.
Other configurations are summarized in Table~\ref{tab:param_highway}.

For collecting a dataset, CEM under the true world model with the true state (i.e. position and velocity of each car) is utilized.
Using it, 52,365 tuples for the training, 971 tuples for the validation, and 2,863 tuples for the test are collected.
Note that noise $\epsilon \sim \mathcal{N}(0, I)$ is injected to the action from CEM for generating various data.

\subsection{Configurations for experiment}

\begin{table}[tb]
    \caption{Configurations for \textit{stretch-reach} and CEM}
    \label{tab:param_stretch}
    \centering
    \begin{tabular}{ccc}
        \hline\hline
        Symbol & Meaning & Value
        \\
        \hline
        $|\mathcal{A}|$ & The size of action space & 2
        \\
        $|\mathcal{X}_{\mathrm{img}}|$ & The size of image observation & 64$\times$64$\times$3
        \\
        $|\mathcal{X}_{\mathrm{arm}}|$ & The size of arm observation & 4
        \\
        -- & Control frequency & 1~Hz
        \\
        $H$ & Horizon step & 5
        \\
        $K$ & The number of candidates & 10,000
        \\
        -- & Maximum iteration & 20
        \\
        $\nu$ & Elite ratio & 0.01
        \\
        $\eta$ & Smooth update & 0.4
        \\
        \hline\hline
    \end{tabular}
\end{table}

The network architecture for VAE is almost identical to Fig.~\ref{fig:impl_vae} and Table~\ref{tab:net_vae} except for replacing $x_{\mathrm{vel}} \to x_{\mathrm{arm}}$ and the latent dimension size $|\mathcal{Z}|=30$.
In contrast, the network architecture for the world model is lightened due to the simplicity of task and importance of real-time control, as shown in Table~\ref{tab:net_world}.

For the control by CEM, the control frequency is set to 1~Hz.
However, under the consideration of other processes, such as encoding the observation, the allowable computational time for CEM is limited to 0.56~s.
Since future predictions do not play a significant role in control performance, $H=5$ is set to reduce the computational cost.
Other configurations are summarized in Table~\ref{tab:param_stretch}.

For collecting a dataset, a simple feedback controller is employed with the explicit target position.
Using it, 2,983 tuples for the training, 63 tuples for the validation, and 107 tuples for the test are collected.
Note that the number of data is absolutely less than one for the simulation due to the lack of diversity.
Therefore, noise injected to the action is also reduced to $\epsilon \sim \mathcal{N}(0, 0.025^2I)$.

\end{document}